%% file: main.tex
\documentclass[sigconf,screen]{acmart}

\usepackage{amsfonts}
\usepackage{algorithmic}
\usepackage{algorithm}
\usepackage{array}
\usepackage{textcomp}
\usepackage{stfloats}
\usepackage{url}
\usepackage{verbatim}
\usepackage{graphicx}
\usepackage{textcomp}
\usepackage{xcolor}
\usepackage{multirow}
\usepackage{siunitx}
\usepackage{comment}
\usepackage{url}
\usepackage{bm}
\usepackage{wrapfig}
\usepackage{enumitem}
\usepackage{cleveref}

\setlist{noitemsep, leftmargin=*, topsep=0pt, partopsep=0pt}
\crefformat{section}{\S#2#1#3} 
\crefformat{subsection}{\S#2#1#3}
\crefformat{subsubsection}{\S#2#1#3}

\newcommand{\SYSTEM}{I-HOPE}


\copyrightyear{2025}
\acmYear{2025}
\setcopyright{cc}
\setcctype{by}
\acmConference[CHASE '25]{ACM/IEEE International Conference on Connected Health: Applications, Systems and Engineering Technologies}{June 24--26, 2025}{New York, NY, USA}
\acmBooktitle{ACM/IEEE International Conference on Connected Health: Applications, Systems and Engineering Technologies (CHASE '25), June 24--26, 2025, New York, NY, USA}
\acmDOI{10.1145/3721201.3721372}
\acmISBN{979-8-4007-1539-6/2025/06}


\begin{document}

\title{Predicting and Understanding College Student Mental Health with Interpretable Machine Learning}


\author{Meghna Roy Chowdhury}
\affiliation{%
   \institution{Purdue University}
   \city{West Lafayette}
   \state{IN}
   \country{USA}}
\email{mroycho@purdue.edu}

\author{Wei Xuan}
\affiliation{%
   \institution{University of Southern California}
   \city{Los Angeles}
   \state{CA}
   \country{USA}}
\email{wxuan@usc.edu}

\author{Shreyas Sen}
\affiliation{%
   \institution{Purdue University}
   \city{West Lafayette}
   \state{IN}
   \country{USA}}
\email{shreyas@purdue.edu}

\author{Yixue Zhao}
\affiliation{%
   \institution{USC Information Sciences Institute}
   \city{Arlington}
   \state{VA}
   \country{USA}}
\email{yzhao@isi.edu}

\author{Yi Ding}
\affiliation{%
   \institution{Purdue University}
   \city{West Lafayette}
   \state{IN}
   \country{USA}}
\email{yiding@purdue.edu}


\begin{abstract}

Mental health issues among college students have reached critical levels, affecting both academic performance and overall wellbeing. Predicting and understanding mental health status among college students is challenging due to three key barriers: the lack of large-scale longitudinal datasets, the prevalence of black-box machine learning models that offer little transparency, and a reliance on population-level analysis rather than personalized understanding.  

To tackle these challenges, this paper presents \SYSTEM{}, the first \textbf{\underline{I}}nterpretable \textbf{\underline{H}}ierarchical m\textbf{\underline{O}}del for \textbf{\underline{P}}ersonalized m\textbf{\underline{E}}ntal health prediction. \SYSTEM{} is a two-stage hierarchical model that connects raw behavioral features to mental health status through five defined behavioral categories as \emph{interaction labels}. We evaluate \SYSTEM{} on the College Experience Study, the longest longitudinal mobile sensing dataset. This dataset spans five years and captures data from both pre-pandemic periods and the COVID-19 pandemic. \SYSTEM{} achieves a prediction accuracy of 91\%, significantly surpassing the 60-70\% accuracy of baseline methods. In addition, \SYSTEM{} distills complex patterns into interpretable and individualized insights, enabling the future development of tailored interventions and improving mental health support. 

\end{abstract}

\maketitle

\input{intro}

\input{dataset}

\input{motivation}
\input{method}
\input{evaluation}
\input{conclusion}

\bibliographystyle{ACM-Reference-Format}
\bibliography{reference}

\appendix

\end{document}

%% file: intro.tex
\section{Introduction}

Mental health issues among college students have escalated to critical levels, significantly affecting academic performance, social interactions, and overall wellbeing~\cite{hammoudi2023understanding,chu2023association,barbayannis2022academic,colarossi2022mental}. The American College Health Association reports that 40\% of students experience severe depression that disrupts daily functioning, while 60\% encounter overwhelming anxiety during the 2020–2021 school year~\cite{abrams2022student}. Furthermore, approximately 76\% of college students report moderate to severe psychological distress, with anxiety and depression being the most prevalent diagnoses~\cite{bestcolleges2024}. Despite increasing awareness, timely access to support remains limited for many students due to stigma, resource constraints, and challenges in detecting those at risk~\cite{Theoverw93:online}. This situation underscores the pressing need for effective and scalable solutions to improve the understanding and prediction of mental health outcomes.

The complexity of predicting and understanding mental health status among college students arises from three primary factors. First, a comprehensive analysis requires a large-scale, longitudinal dataset that collects data through passive sensing over an extended period rather than relying on short-term data collection conducted in a lab setting. Second, although machine learning has been used to address mental health issues, many existing models utilize black-box algorithms that lack transparency and interpretability~\cite{xu2024medical,sajno2023machine,tutun2023ai}. Third, most machine learning approaches yield aggregated insights at the population level that fail to provide individualized understanding, which is essential for personalized interventions and mental health support~\cite{batstra2024more,Findingp69:online}.

In this paper, we address these challenges through a paradigm shift towards methodologies that not only leverage an extensive dataset but also prioritize individual variability in mental health prediction. In particular, we leverage the College Experience Study (CES) dataset~\cite{nepal2024capturing}, the longest longitudinal mobile sensing dataset for college student behaviors, released by Dartmouth College in October 2024. This dataset is especially valuable for our study because it covers a five-year period that includes pre-pandemic years, the COVID-19 pandemic, and the gradual return to normalcy as the pandemic receded. By analyzing behavioral patterns and mental health metrics over different time periods, we can investigate insights into the pandemic's impact on mental health and the role of behavior in shaping it.

We present \SYSTEM{}, the first \textbf{\underline{I}}nterpretable \textbf{\underline{H}}ierarchical m\textbf{\underline{O}}del for \textbf{\underline{P}}ersonalized m\textbf{\underline{E}}ntal health prediction. \SYSTEM{} is a two-stage hierarchical model designed to accurately predict mental health status while offering deep insights into the features contributing to various mental health conditions. Specifically, we define five interaction labels: \emph{Leisure}, \emph{Me Time}, \emph{Phone Time}, \emph{Sleep}, and \emph{Social Time}, which categorize different daily behaviors. The key insight of \SYSTEM{} is to connect raw behavioral features to mental health status through an intermediate layer consisting of these five interaction labels. These labels act as compact representations of complex behaviors, simplifying the input space while preserving data richness. Furthermore, \SYSTEM{} facilitates the identification of an individual's social type and emphasizes the interactions that most significantly impact their mental health.

We compare \SYSTEM{} with baseline methods that lack personalized predictions or two-stage feature mappings. \SYSTEM{} achieves an overall prediction accuracy of 91\%, significantly surpassing the 60-70\% accuracy of baseline methods. \SYSTEM{} simplifies complex patterns into interpretable and personalized insights by mapping behavioral features into interaction labels. For instance, a good 7-hour sleep is linked to better mental health outcomes, while poor sleep correlates with increased anxiety and depression. Walking serves as a stress reliever associated with leisure and relaxation. Phone usage reveals contrasting patterns; interactions at home often indicate positive connections, whereas excessive use in social settings suggests stress or discomfort. The balance between social and personal time is crucial, with shared spaces that promote engagement and personal spaces that allow emotional recharge. Importantly, these behaviors affect individuals differently; some find ``Me Time'' in working out, while others prefer studying or spending time in their own dorms.  The code is available at \url{https://github.com/roycmeghna/I-HOPE}.

We summarize the contributions as follows.
\begin{itemize}
    \item \textbf{Interpretable machine learning framework development:} We present \SYSTEM{}, the first hierarchical model that maps raw behavioral data into five interpretable interaction labels: \emph{Leisure}, \emph{Me Time}, \emph{Phone Time}, \emph{Sleep}, and \emph{Social Time}, enhancing the transparency and prediction accuracy.
    \item \textbf{Personalized mental health predictions:} \SYSTEM{} adapts to individual behaviors, enhancing the accuracy and relevance of mental health assessments.
    \item \textbf{Key behavioral predictors identification:} \SYSTEM{} identifies specific behaviors, such as sleep patterns and physical activity levels, significantly influencing mental health outcomes, providing insight into targeted interventions.
    \item \textbf{Scalable analytical approach:} \SYSTEM{} scales high-dimensional behavioral data analysis, linking complex datasets to real-world mental health applications.
\end{itemize}

%% file: dataset.tex
\section{College Experience Study (CES) Dataset}\label{sec_dataset} 

To thoroughly investigate college students' mental health status, a large-scale dataset that captures daily behaviors is required. Therefore, we leverage the CES dataset~\cite{nepal2024capturing}, the longest longitudinal mobile sensing dataset for college student behaviors, released by Dartmouth College in October 2024. This dataset includes passive mobile sensing data--- mobility, physical activity, sleep patterns, and phone usage---along with Ecological Momentary Assessment (EMA) surveys from 217 Dartmouth students collected between 2017 and 2022. It comprises over 210,000 data points collected across two cohorts throughout their college years on an hourly basis. The EMA surveys are delivered randomly once a week via the StudentLife mobile application~\cite{wang2014studentlife}.

This dataset is especially valuable for our study because it covers a five-year period that includes pre-pandemic years, the COVID-19 pandemic, and the gradual return to normalcy as the pandemic receded. By analyzing behavioral patterns and mental health metrics over different time periods, we can assess and predict how mobile sensing data—collected before, during, and after the COVID-19 pandemic—predicts students' mental health. This analysis provides valuable insights into the pandemic's impact on mental health and the role of behavior in shaping it.

\begin{table}[!t]
\centering
\caption{\textbf{PHQ-4 scores and their categories.}}
\begin{tabular}{l|l} \toprule
\textbf{PHQ-4 score}& \textbf{Category}\\ \midrule
 0-3& Normal\\ \hline 
 4-6& Mild\\\hline
 7-9& Moderate\\\hline
 10-12& Severe\\\bottomrule 
 \end{tabular} 
    \label{tbl:phq4}
\end{table}

\begin{figure}[t]
    \centering
    \includegraphics[width=\linewidth]{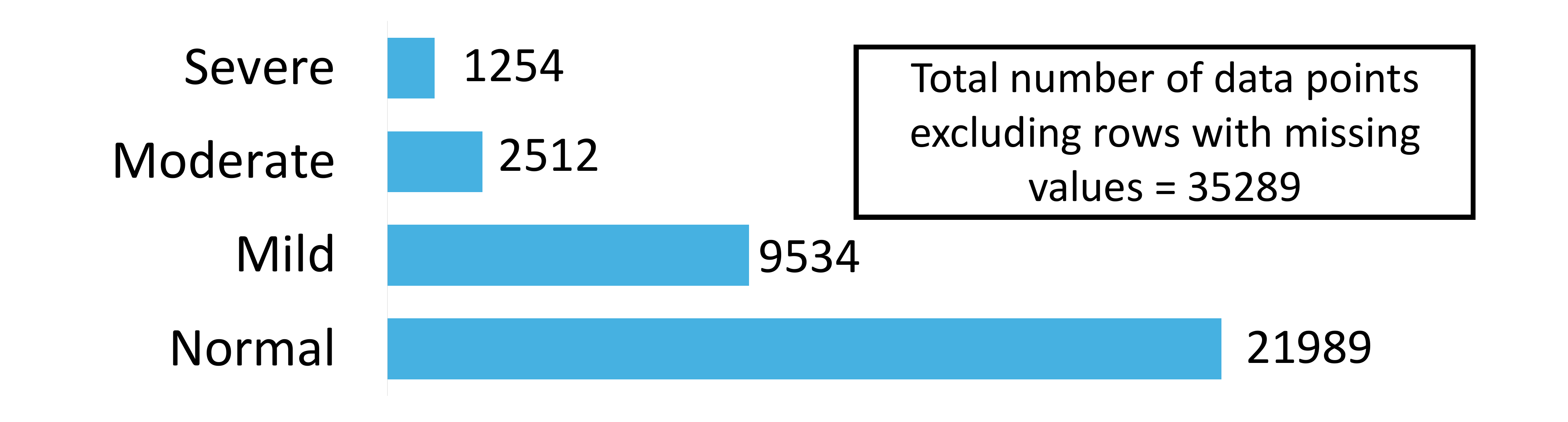}
    \caption{\textbf{Data distribution across PHQ-4 categories}. Normal has the highest count, whereas Severe has the lowest count.}
    \label{fig_datadist}
\end{figure}

\paragraph{PHQ-4 score.} The EMA survey includes a key mental health metric, Patient Health Questionnaire-4 (\emph{PHQ-4}) score~\cite{kroenke2009ultra}, which serves as the focal point for our analysis and the mental health outcome we aim to predict in this paper. The PHQ-4 score is a widely recognized screening tool for assessing depressive and anxiety symptoms in clinical settings and epidemiological studies~\cite{lowe20104}. It ranges from 0 to 12, with lower scores indicating better mental health. To enhance interpretability and align with standardization practices~\cite{wicke2022update}, we categorize PHQ-4 scores into four levels: \emph{normal}, \emph{mild}, \emph{moderate}, and \emph{severe}. Each category corresponds to a specific PHQ-4 score range as shown in~\Cref{tbl:phq4}, which allows clearer insight into mental health status. The distribution of data points across the PHQ-4 categories is shown in~\Cref{fig_datadist}.

\begin{table}[!t]
\centering
\caption{\textbf{Pearson correlations between PHQ-4 score with smartphone usage features in five environments.} } 
\begin{tabular}{l|l|l}
\toprule
\textbf{Category}   & \textbf{Metric}             & \textbf{Correlation} \\ \midrule
\multirow{2}{*}{Overall}   & Duration of Unlock  & 0.0321 \\ \cline{2-3} 
                                    & \# Unlocks         & -0.0107   \\ \hline
                                    
\multirow{2}{*}{Food}      & Duration of Unlock  & -0.0125 \\ \cline{2-3} 
                                    & \# Unlocks         & -0.0197 \\ \hline
\multirow{2}{*}{Study}     & Duration of Unlock  & 0.0014  \\ \cline{2-3} 
                                    & \# Unlocks         & -0.0151 \\ \hline
\multirow{2}{*}{Social}    & Duration of Unlock  & -0.01   \\ \cline{2-3} 
                                    & \# Unlocks         & -0.0157 \\ \hline
\multirow{2}{*}{Dormitory} & Duration of Unlock  & -0.00215 \\ \cline{2-3} 
                                    & \# Unlocks         & -0.01505 \\ \hline
\multirow{2}{*}{Home}      & Duration of Unlock  & 0.0188  \\ \cline{2-3} 
                                    & \# Unlocks         & 0.0033  \\ \bottomrule 
\end{tabular}
\label{tbl:corr}
\end{table}

\paragraph{Features.} The original dataset has 172 features, including smartphone usage, duration spent performing various activities, duration spent at various locations, sleep information, etc. However, not all features directly relate to mental health studies, such as the amplitude of detected audio. Additionally, some features, such as distance traveled and number of locations visited, are ambiguous. After careful screening, we select 45 features for our study. These features, which include contextual information, phone usage, and sleep duration, are chosen based on the following criteria. 

\begin{enumerate}
\item They provide a meaningful representation of mental health status by capturing behavioral patterns that research has consistently linked to mental health.
\item They demonstrated statistical significance with respect to the PHQ-4 categories, with p-values less than 0.05 and, in many cases, less than 0.01.
\end{enumerate}

%% file: motivation.tex
\section{Motivation} \label{sec_motivation}

In this section, we explore the challenges of predicting mental health status by analyzing the CES dataset and identify key insights for designing effective predictive models.

\subsection{The Role of Smartphone Usage on Mental Health}

In mental health research, smartphone usage has increasingly been used to facilitate real-time data collection, improve access to mental health resources, and enable personalized interventions. Prior work has shown that screen time and social media use negatively impact mental health and digital wellbeing~\cite{zhao2024digital,elamin2024smartphone}. Mobile sensing and interventions have also shown promise for personalizing mental health support~\cite{sheng_yu_careforme_2024}. Therefore, we start by focusing on two smartphone usage metrics: \emph{the number of times phones are unlocked (\#Unlocks)} and \emph{daily phone usage duration (Duration of Unlock)}.

To explore the relationship between these features and mental health, we conduct a correlation analysis using the CES dataset. The results, summarized in~\Cref{tbl:corr}, indicate three main findings: (i) longer phone usage correlates with poorer mental health; (ii) higher unlock frequencies are associated with better mental health; (iii) increased phone use in social and dining settings relates to worse mental health, possibly due to social anxiety, while greater use at home suggests positive activities like socialization, leisure, or learning. These findings highlight the importance of context in understanding the impact of smartphone usage on mental health.

\subsection{Smartphone Usage Features Alone Are Insufficient for Predicting Mental Health}

To assess whether smartphone usage features can effectively predict mental health outcomes, we train a machine learning model on the CES dataset to predict the PHQ-4 score. Given that the PHQ-4 score has four categories, as illustrated in~\Cref{tbl:phq4}, this constitutes a multi-class classification problem. Therefore, we use two smartphone features mentioned above---the number of unlocks and daily phone usage duration---as input features to train a multilayer perceptron (MLP) model, a neural network widely used for classification and prediction~\cite{popescu2009multilayer}. The MLP model has 2 input nodes, 3 fully connected hidden layers, and 4 output nodes corresponding to the PHQ-4 categories. The model is trained for 50 epochs until the learning curve stabilizes. We preprocess the data by normalizing the features and ensuring balanced class distributions through oversampling for the PHQ-4 categories. We randomly split the dataset into 80\% for training and 20\% for testing. Our results reveal that the prediction accuracy is only 28\%, which is low. These results indicate that relying solely on these smartphone usage features is insufficient to predict mental health outcomes accurately. 

\subsection{Improving Predictions with Additional Features}

To improve prediction accuracy, we need to incorporate additional features from the CES dataset. Specifically, we use 45 manually selected features from the CES dataset related to location information, sleep duration, and activity times (see Column 2 of \Cref{tbl:interaction-label}). We use the MLP model with 45 input nodes, 3 fully connected hidden layers, and 4 output nodes corresponding to the PHQ-4 categories. The training and test sets are the same as above. The model is trained for 50 epochs until the learning curve stabilizes. This approach improves the prediction accuracy to 60\%, which is still not high. We attribute the low accuracy to two reasons: (i) treating the dataset as a whole may have obscured individual differences; (ii) a high correlation among features likely reduce the model's ability to make accurate predictions. We discuss these below.

\paragraph{Addressing Individual Differences}

The CES dataset includes data from 217 individuals, each contributing about 160 data points on average. Research indicates that individual differences in behavior and preferences can significantly affect predictive modeling outcomes~\cite{he2023personalized,lengerich2019learning}. To capture these variations, we develop personalized MLP model for each student. We focus on students who contribute to at least 160 data points in the dataset, which gives us 121 students. Each model is trained on the corresponding individual's data with 45 selected features, maintaining the same MLP architecture and train-test split as described above. This personalized approach improves the average prediction accuracy to 70\%. This motivates the need for additional feature engineering to improve feature representation and prediction accuracy.

\paragraph{Reducing Feature Correlation}

\begin{figure}[!t]
    \centering
    \includegraphics[width=0.8\linewidth]{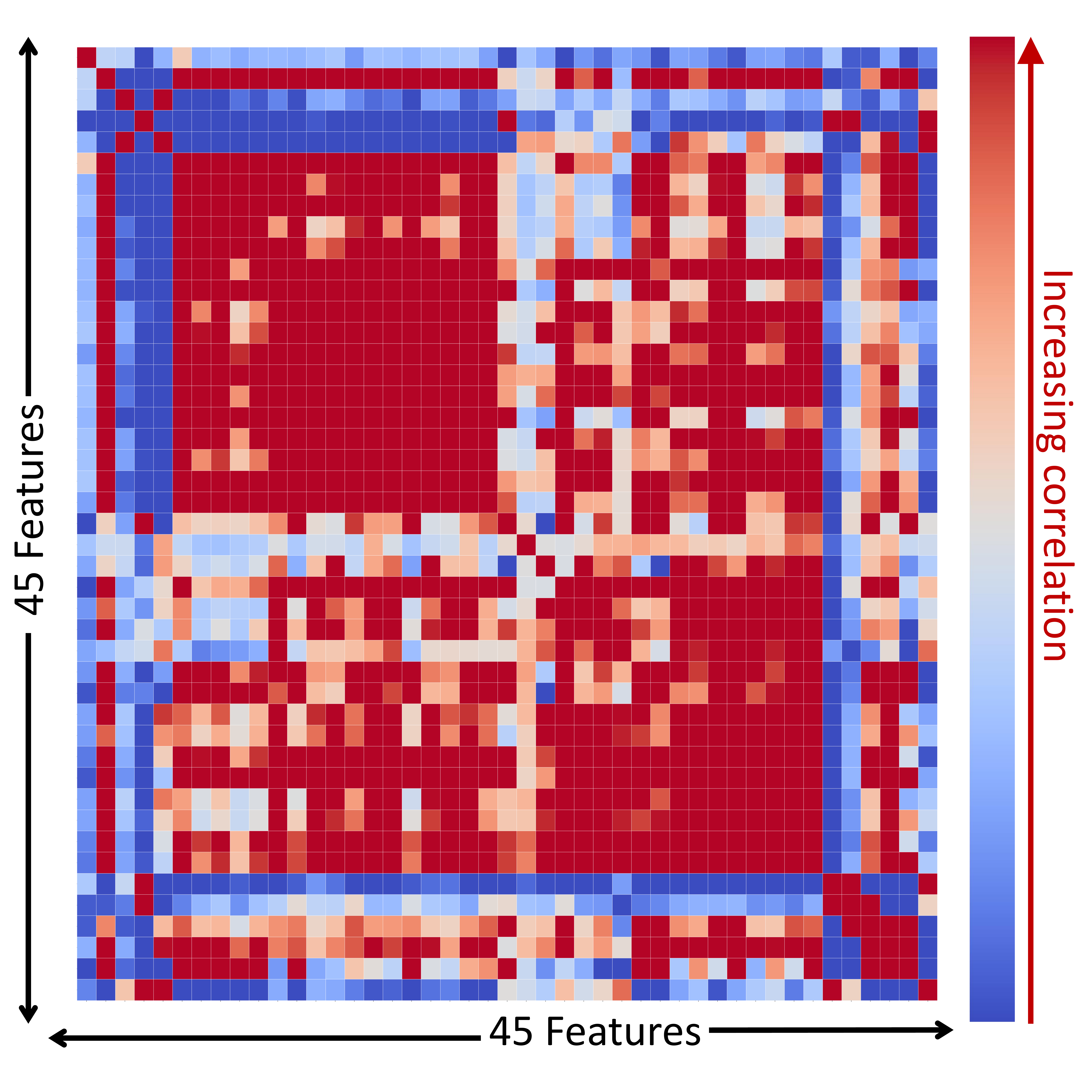}
    \caption{\textbf{Heatmap for correlations among features.}}
    \label{fig_motivation_highcorr}
\end{figure}

\begin{figure}[!t]
    \centering
    \includegraphics[width=0.7\linewidth]{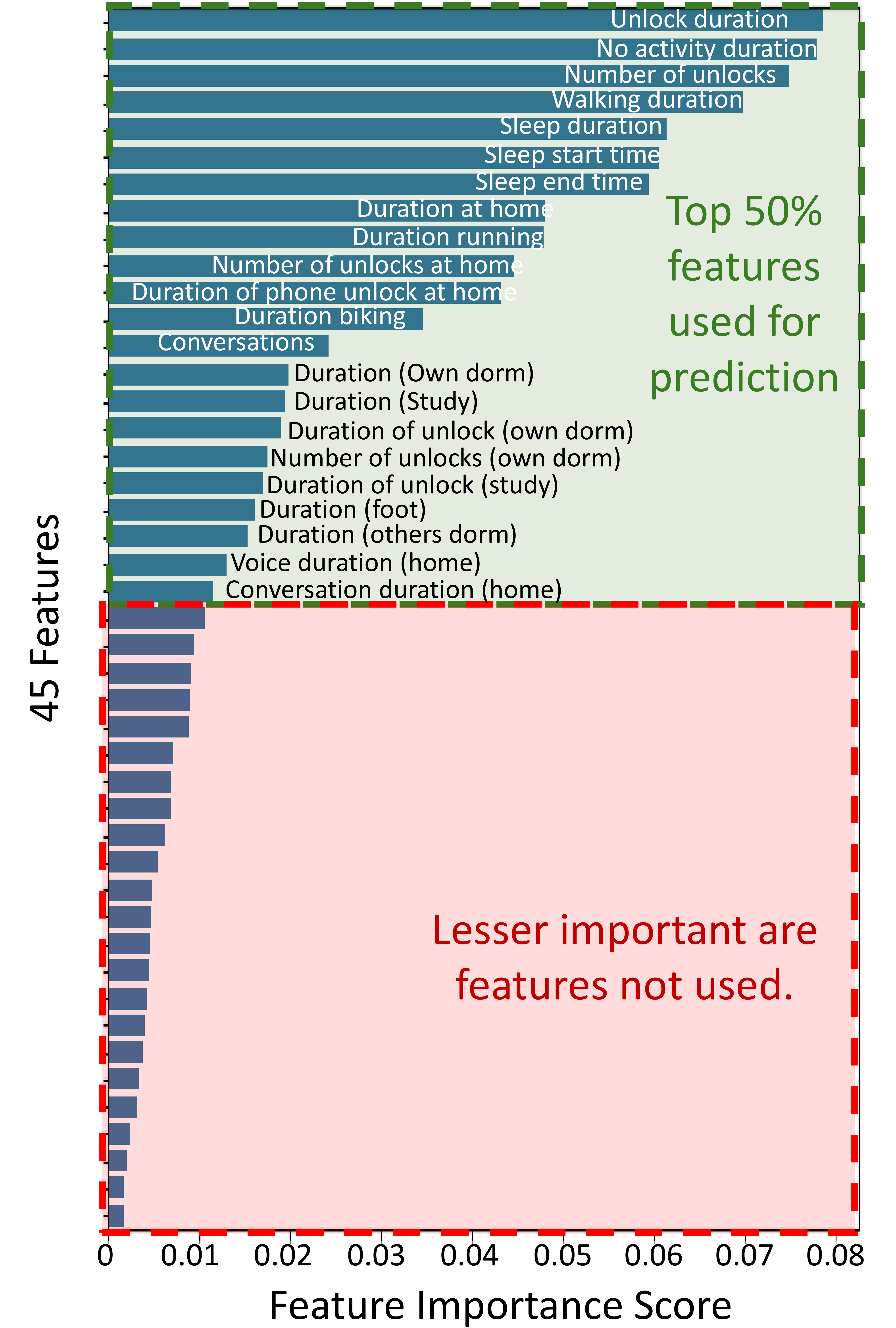}
    \caption{\textbf{Feature importances using random forests.}}
    \label{fig_motivation3_imp}
\end{figure}

Highly correlated features introduce redundancy, leading to overfitting and reduced test accuracy while increasing computational complexity. As shown in \Cref{fig_motivation_highcorr}, many features are highly correlated. To address this, we analyze feature importance using random forest, an efficient and interpretable ensemble learning technique~\cite{breiman2001random}. This method builds multiple decision trees and combines their outputs to evaluate feature importance. \Cref{fig_motivation3_imp} shows the importance ranking of all 45 features, highlighting the top ones. We select the top 50\% of these features to retrain personalized MLP models, adjusting the input dimensions accordingly. However, this results in slight a drop in prediction accuracy by $\sim$5\%, indicating that simply excluding certain features may remove critical information necessary for accurate predictions.

\paragraph{Key takeaway.} These analyses suggest that the features influencing prediction accuracy vary among individuals. For instance, a higher biking time may correlate with better mental health for one person, while walking may be more relevant for another, and a combination of running and walking could be key for someone else. A global feature importance approach risks eliminating crucial features for individuals, potentially reducing accuracy. Therefore, it is vital to incorporate personalization while minimizing feature correlation, ensuring the model captures individual-specific patterns without being hindered by redundant or irrelevant features, ultimately enhancing overall prediction accuracy.

%% file: method.tex
\begin{figure*}[!t]
    \centering
    \includegraphics[width=0.75\linewidth]{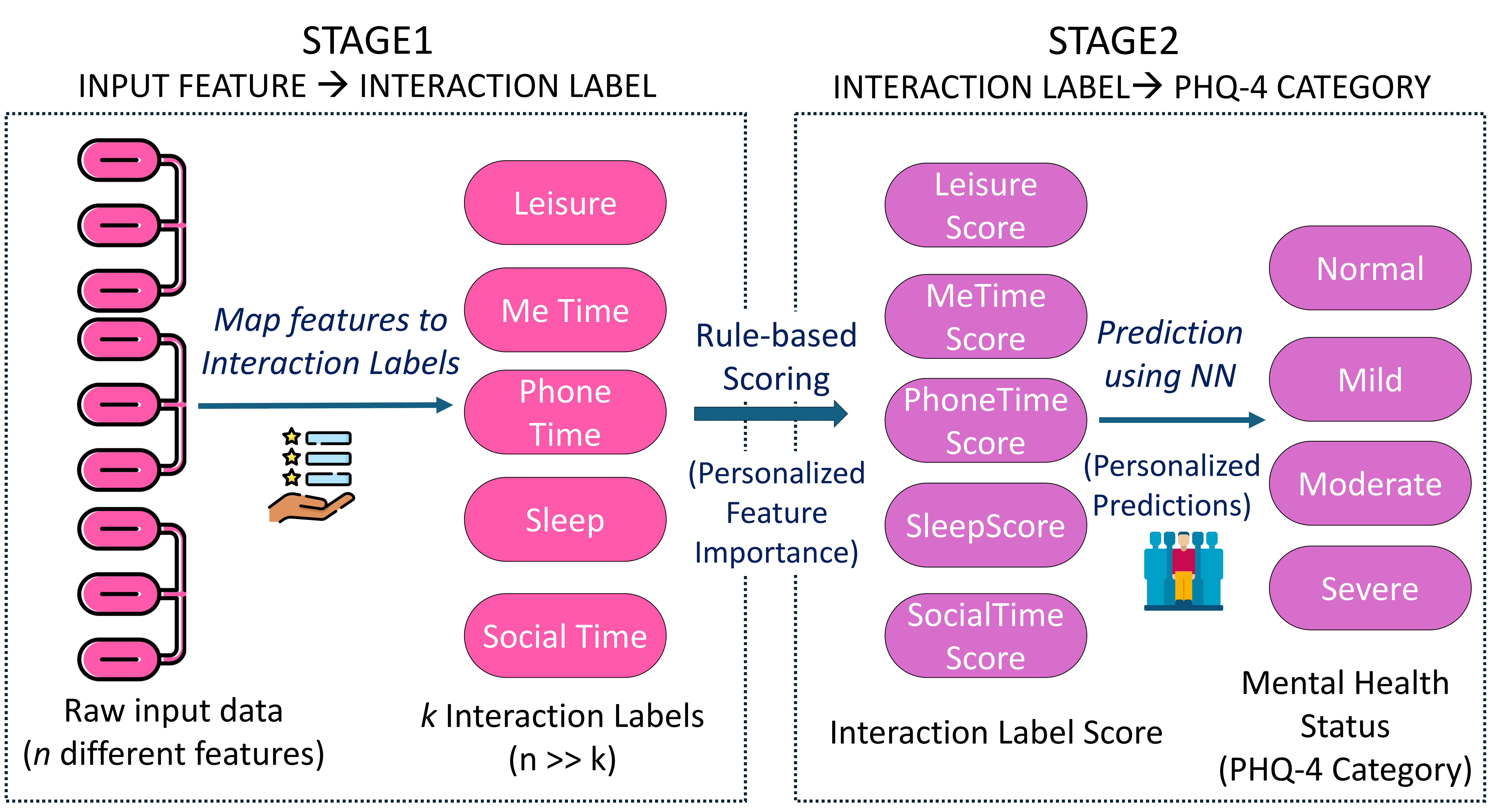}
    \caption{\textbf{Workflow of the \SYSTEM{} design.}}
    \label{fig_workflow} 
\end{figure*}

\section{\SYSTEM{}}


Our analysis has shown that (1) personalized models outperform a global model shared among students; and (2) effective feature engineering efforts are needed to reduce feature correlation and improve prediction accuracy. Therefore, we design \SYSTEM{}, the first \textbf{\underline{I}}nterpretable \textbf{\underline{H}}ierarchical m\textbf{\underline{O}}del for \textbf{\underline{P}}ersonalized m\textbf{\underline{E}}ntal health prediction. \SYSTEM{} is a two-stage hierarchical model that connects behavioral features to mental health status through an intermediate layer of five interaction labels: \emph{Leisure}, \emph{Me Time}, \emph{Phone Time}, \emph{Sleep}, and \emph{Social Time}, which categorize different daily behaviors. This model structure offers two benefits:

\begin{enumerate}
    \item \textbf{Dimension Reduction.} The interaction labels serve as compact representations of complex behaviors, simplifying the input space while retaining the richness of the data. The effective feature representation contributes to high prediction accuracy.
    \item \textbf{Enhanced Interpretability.} \SYSTEM{} enables the identification of an individual's social type and highlights the interactions that correlate their mental health the most.
\end{enumerate}




\Cref{fig_workflow} presents the workflow of \SYSTEM{} design. It starts with \emph{n} input features, which are mapped to \emph{k} interaction labels ($k=5$), representing aggregated behavioral categories: \emph{Leisure}, \emph{Me Time}, \emph{Phone Time}, \emph{Sleep}, and \emph{Social Time}. To quantify the behavioral insights, we assign a score for each interaction label by introducing a two-step rule-based method.  These scores then serve as input to a neural network model that predicts one of four mental health categories: \emph{Normal}, \emph{Mild}, \emph{Moderate}, or \emph{Severe}. In summary, by leveraging the hierarchical model structure, \SYSTEM{} improves both prediction accuracy and model interpretability. For example, \SYSTEM{} can assess whether ``walking'' or ``spending time with friends'' in the \emph{Leisure} category has a greater impact on mental health of an individual. \SYSTEM{} is conducted per-user basis by training and testing on individualized data. Before introducing each stage in detail, we will first  describe the feature engineering efforts we have made to prepare the final input dataset for \SYSTEM{}.

\begin{table*}[!t]
    \centering
        \caption{\textbf{Interaction labels and their corresponding features.}}
    \renewcommand{\arraystretch}{1} 
    \setlength{\tabcolsep}{2pt} 
    \small 
    \begin{tabular}{>{\centering\arraybackslash}m{2.6cm}|>{\arraybackslash}m{10cm}} \toprule
         \textbf{Interaction label} & \textbf{Relevant features corresponding to each interaction label} \\ \hline
         Leisure  & 
         \parbox{12cm}{
         \begin{itemize}\setlength{\itemsep}{0pt}\setlength{\parskip}{0pt}
         \vspace{1mm}
            \item Duration of physical activities (biking, walking, running)
            \item Duration of conversations (in-person and phone)
            \item Duration in locations like others' dorm, workout
            \item Duration of phone usage in various locations
            \vspace{1mm}
         \end{itemize}
         } \\ \hline 
         Me Time  & 
         \parbox{12cm}{
         \begin{itemize}\setlength{\itemsep}{0pt}\setlength{\parskip}{0pt}
         \vspace{1mm}
            \item Duration of activities (biking, walking, running, being still, studying)
            \item Various measures at home (conversation detected, duration)
            \item Conversations detected at own dorm or home
            \item Duration of phone usage in various locations
            \vspace{1mm}
         \end{itemize}
         } \\ \hline
         Phone Time  & 
         \parbox{12cm}{
         \begin{itemize}\setlength{\itemsep}{0pt}\setlength{\parskip}{0pt}
         \vspace{1mm}
            \item Duration of phone conversations in various locations
            \item Ratio of number of calls to duration of calls
            \item Ratio of number of phone unlocks to duration of unlocks, in various locations
            \vspace{1mm}
         \end{itemize}
         } \\ \hline
        Sleep  & 
        \parbox{12cm}{
        \begin{itemize}\setlength{\itemsep}{0pt}\setlength{\parskip}{0pt}
        \vspace{1mm}
            \item Duration of stillness
            \item Audio detection at home or own dorm
            \item Phone usage at night
            \item Sleep duration
            \vspace{1mm}
        \end{itemize}
        } \\ \hline
        Social Time  & 
        \parbox{12cm}{
        \begin{itemize}\setlength{\itemsep}{0pt}\setlength{\parskip}{0pt}
        \vspace{1mm}
            \item Duration of social activities (biking, walking, running, workout, study, eating food)
            \item Ratio of number of calls to duration of calls at locations like study space, home, etc.
            \item Amount of time spent in others' dorm
            \vspace{1mm}
        \end{itemize}
        } \\ \bottomrule
    \end{tabular}
\label{tbl:interaction-label}
\end{table*}

\subsection{Feature Engineering}


In \Cref{sec_motivation}, we have demonstrated that high quality input features are critical to high prediction accuracy. Based on our previous analysis, we further conduct the following feature engineering procedure to prepare the input dataset for \SYSTEM{}.

In our case, feature engineering is guided by domain relevance, ensuring that derived features carry meaningful interpretations in the context of mental health. For example, rather than using the number of phone unlocks and duration of phone usage as separate features, we introduce a ratio feature: \texttt{number\_of\_unlocks / duration\_of\_usage}. A high value of this feature suggests more frequent, brief interactions with the phone, and potentially indicative of restlessness. We apply similar transformation across multiple contextual locations, including home, own dorm, study spaces, others' dorms, social environments, and overall daily usage. In addition, instead of separately including the number and duration of incoming and outgoing calls, we combine them into a single feature:\[
\frac{\texttt{Number of incoming calls + outgoing calls}}{\texttt{Call duration of incoming + outgoing calls}}.
\] 
This consolidation maintains the behavioral signal while reducing redundancy. Overall, we reduce the 45 features to 35 final features, preserving domain-specific behavioral meaning while improving computational efficiency.

\subsection{Stage 1: Input Features $\rightarrow$ Interaction Labels}

In Stage 1, \SYSTEM{} maps input 35 features to five interaction labels: \emph{Leisure}, \emph{Me Time}, \emph{Phone Time}, \emph{Sleep}, and \emph{Social Time}. This approach combines domain knowledge and data-driven insights to capture relevant behavioral dimensions, reduce redundancy, and enhance interpretability while maintaining granularity for personalized analysis. The detailed process is described below.

\begin{algorithm}[t]
\caption{Rule-based initialization of \emph{SleepScore}}
\begin{algorithmic}[1]
    \STATE \textbf{Input:} Thresholds of each feature in \emph{Sleep} label, which are population means of the feature distributions.
    \STATE \textbf{Initialize:} SleepScore = 0
    \IF{\texttt{Sleep duration} $>$ Threshold  (\textbf{where} Threshold = Mean of \texttt{Sleep duration} distribution)}
        \STATE SleepScore += 1
    \ENDIF
    \IF{\texttt{Duration at own dorm} $>$ Threshold  (\textbf{where} Threshold = Mean of \texttt{Duration at own dorm} distribution)}
        \STATE SleepScore += 1
    \ENDIF
    \IF{\texttt{Duration of stillness} $>$ Threshold  (\textbf{where} Threshold = Mean of \texttt{Duration of stillness} distribution)}
        \STATE SleepScore += 1
    \ENDIF
    \IF{\texttt{Conversation detected at home} $<$ Threshold  (\textbf{where} Threshold = Mean of \texttt{Conversation detected at home} distribution)}
        \STATE SleepScore += 1
    \ENDIF
    \STATE \textbf{Output:} SleepScore
\end{algorithmic}\label{alg:sleep-init}
\end{algorithm}

\begin{enumerate}
    \item \textbf{Initial mapping:} We manually map each feature to different each interaction label based on their semantic and behavioral relevance. For instance, features like \emph{unlock duration} and \emph{phone conversations} belong to \emph{Phone Time}, while \emph{duration of social activities} and \emph{time spent with friends} belong to \emph{Social Time}. The initial mapping is detailed in \Cref{tbl:interaction-label}. Notably, some features are assigned to multiple labels, reflecting their multidimensional impact on various behavioral categories. We further employ K-means clustering with k=5 to perform a soft validation, assessing whether five interaction labels are a reasonable choice.
    \item \textbf{Rule-based scoring initialization:} Since interaction labels are manually created, we need to assign them numerical values for downstreaming quantitative analysis. \SYSTEM{} uses a two-step rule-based scoring method to compute an \emph{[InteractionLabel]Score} for each label. The first step is to initialize the scores. Specifically, for each feature within a label, we compare its value to the population mean value of that feature; if it exceeds the mean, the score increases by 1. The intuition is that higher feature values suggest stronger relevance to the label and thus contribute more to its score. We illustrate the first step of this method using the \emph{Sleep} label in \Cref{alg:sleep-init}, although the same procedure applies to all interaction labels. 
    \item \textbf{Rule-based scoring using feature importance:} The second step of the rule-based scoring method is to use a random forest model on the features and initialized scores to assess the contribution of individual input features to each interaction label. We choose random forest due to its ability to handle high-dimensional data and capture nonlinear relationships~\cite{breiman2001random}. \SYSTEM{} then uses feature importance generated from random forest to refine the final scores. For each feature within a label, we compare its value to the population mean; if its value exceeds the mean, we increase the score by its Normalized Weighted Feature Impact (NWFI), which is a weighted measure combining the feature's random forest-derived importance and its normalized value. The normalization reflects the scaling of feature values, and the weighted indicates the incorporation of random forest feature importance. We illustrate the second step of this method using the \emph{Sleep} label in~\Cref{alg:sleep-final}, although the same procedure applies to all interaction labels. 
\end{enumerate}

\begin{algorithm}[t]
\caption{Rule-based final scoring of \emph{SleepScore}}
\begin{algorithmic}[1]
    \STATE \textbf{Input:} Thresholds of each feature in \emph{Sleep} label, which are population means of the feature distributions.
    \STATE \textbf{Initialize:} SleepScore = 0
    \IF{\texttt{Sleep duration} $>$ Threshold  (\textbf{where} Threshold = Mean of \texttt{Sleep duration} distribution)}
        \STATE SleepScore += NWFI of \texttt{Sleep duration}
    \ENDIF
    \IF{\texttt{Duration at own dorm} $>$ Threshold  (\textbf{where} Threshold = Mean of \texttt{Duration at own dorm} distribution)}
        \STATE SleepScore += NWFI of \texttt{Duration at own dorm}
    \ENDIF
    \IF{\texttt{Duration of stillness} $>$ Threshold  (\textbf{where} Threshold = Mean of \texttt{Duration of stillness} distribution)}
        \STATE SleepScore += NWFI of \texttt{Duration of stillness}
    \ENDIF
    \IF{\texttt{Conversation detected at home} $<$ Threshold  (\textbf{where} Threshold = Mean of \texttt{Conversation detected at home} distribution)}
        \STATE SleepScore += NWFI of \texttt{Conversation detected at home}
    \ENDIF
    \STATE \textbf{Output:} SleepScore
\end{algorithmic}\label{alg:sleep-final}
\end{algorithm}

\subsection{Stage 2: Interaction Labels $\rightarrow$ PHQ-4 Category}

In Stage 2, \SYSTEM{} uses the five interaction label scores (\emph{LeisureScore}, \emph{MeScore}, \emph{PhoneScore}, \emph{SleepScore}, \emph{SocialScore}) as inputs to predict PHQ-4 categories. This stage simplifies the predictive process by using these labels as compact and interpretable representations of the raw data. Specifically, \SYSTEM{} builds a neural network model to predict PHQ-4 categories using the five interaction label scores as inputs. The neural network includes an input layer with five nodes, three hidden layers for pattern extraction, and an output layer with four nodes for the PHQ-4 categories (\emph{Normal}, \emph{Mild}, \emph{Moderate}, or \emph{Severe}). \SYSTEM{} uses the ReLU activation function for hidden layers and the Softmax activation function for the output layer for multi-class classification. The neural network model is trained for 50 epochs using the Adam optimizer~\cite{kingma2014adam} with a learning rate of 0.001 and categorical cross-entropy as the loss function.

\subsection{Implementation}

We implement \SYSTEM{} in Python. We use \texttt{Pandas} and \texttt{NumPy} for data preprocessing (cleaning, transformation, and feature extraction) and feature engineering. We implement neural network models using \texttt{TensorFlow}. Evaluation metrics such as accuracy are computed with \texttt{Scikit-learn}.

%% file: evaluation.tex
\section{Evaluation}

In this section, we evaluate the \SYSTEM{} to address the following research questions (RQs):

\begin{enumerate}[label=\textbf{RQ\arabic*.}]
    \item \textbf{Prediction accuracy:} How accurately does \SYSTEM{} predict the mental health status of college students?  
    \item \textbf{Interpretability from hierarchical mapping:} What new insights into mental health can we gain from hierarchical feature mapping of \SYSTEM{}? 
    \item \textbf{Scoring for interaction labels:} How much do features contribute to the scoring of interaction labels?  
\end{enumerate}

The rest of this section is divided into three parts, each addressing a research question.

\subsection{Prediction Accuracy}
\begin{table*}[!t]
    \centering
    \caption{\textbf{Comparison of different methods for predicting PHQ-4 categories.} Evaluation metrics include precision (Pre.), recall (Rec.), F1-score (F1), and overall accuracy for each PHQ-4 category.}
    \begin{tabular}{l|lll|lll|lll|lll} \toprule
         \multirow{2}{*}{\textbf{PHQ-4 Category}} 
         & \multicolumn{3}{c|}{\textbf{Baseline 1}} 
         & \multicolumn{3}{c|}{\textbf{Baseline 2}} 
         & \multicolumn{3}{c|}{\textbf{Baseline 3}} 
         & \multicolumn{3}{c}{\textbf{\SYSTEM{}}} \\  
         \cmidrule(lr){2-4} \cmidrule(lr){5-7} \cmidrule(lr){8-10} \cmidrule(lr){11-13}
         & \textbf{Prec.} & \textbf{Rec.} & \textbf{F1}  
         & \textbf{Prec.} & \textbf{Rec.} & \textbf{F1}  
         & \textbf{Prec.} & \textbf{Rec.} & \textbf{F1}
         & \textbf{Prec.} & \textbf{Rec.} & \textbf{F1} \\ \midrule
         \textbf{Normal}  & 0.6453 & 0.627 & 0.6360 & 0.7149 & 0.677 & 0.6954 & 0.6781 & 0.654 & 0.6659 & 0.9513 & 0.9285 & 0.9398 \\ \hline
         \textbf{Mild}    & 0.6135 & 0.617 & 0.6152 & 0.6432 & 0.667 & 0.6549 & 0.6297 & 0.667 & 0.6478 & 0.9489 & 0.9285 & 0.9386 \\ \hline
         \textbf{Moderate} & 0.5908 & 0.600 & 0.5953 & 0.6371 & 0.660 & 0.6483 & 0.6083 & 0.660 & 0.6332 & 0.8791 & 0.8690 & 0.8740 \\ \hline
         \textbf{Severe}  & 0.5831 & 0.559 & 0.5708 & 0.6265 & 0.614 & 0.6202 & 0.5982 & 0.614 & 0.6060 & 0.8645 & 0.9138 & 0.8884 \\ \midrule
         \textbf{Overall Accuracy} & \multicolumn{3}{c|}{0.60} & \multicolumn{3}{c|}{0.70} & \multicolumn{3}{c|}{0.65} & \multicolumn{3}{c}{\textbf{0.91}} \\ \bottomrule         
    \end{tabular}
    \label{table_cm}
\end{table*}
\begin{figure}[t]
    \centering
    \includegraphics[width=0.9\linewidth]{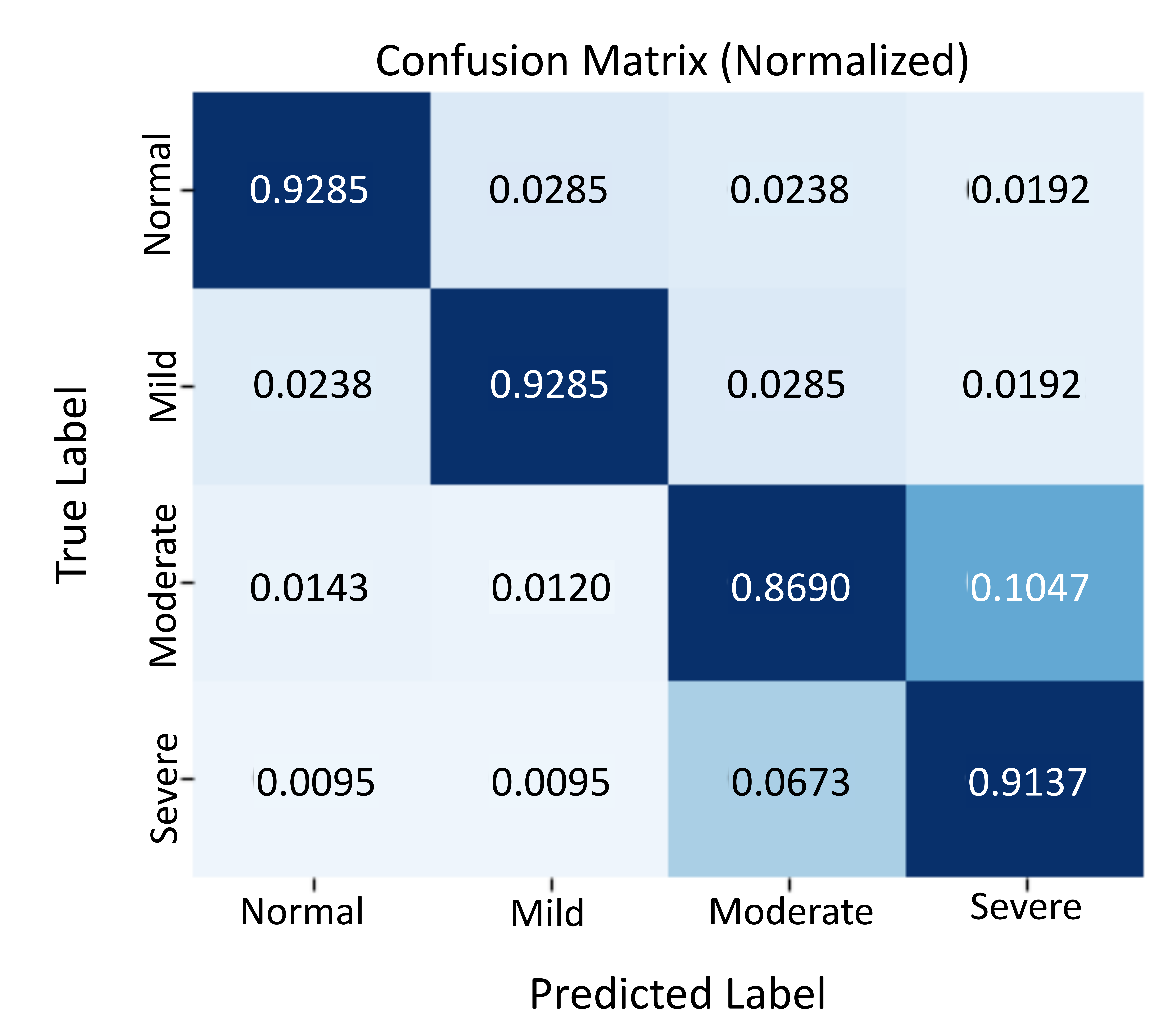}
    \caption{{\textbf{Normalized confusion matrix from \SYSTEM{}.}}}
    \label{fig_cm}
\end{figure}

We compare \SYSTEM{} against three baseline methods that differ in the training approach (aggregated vs. personalized), feature extraction (all raw features vs. selected features) and model structure (single stage vs. hierarchical). We evaluate 121 students who controbut at least 160 data points in the dataset. We report results from the best-performing model from 5-fold cross validation for all baselines. The methods are summarized as follows, all using the previously described neural network architecture with adjusted input dimensions:

\begin{itemize}
    \item \textbf{Baseline 1:} An aggregated single model trained with selected 45 raw features across 121 students.   
    \item \textbf{Baseline 2:} Personalized models for each individual using selected 45 raw features.
    \item \textbf{Baseline 3:} Personalized models for each individual using the top 50\% of 45 features based on feature importance from random forest models. 
    \item \textbf{\SYSTEM{}:} The \textbf{\underline{I}}nterpretable \textbf{\underline{H}}ierarchical m\textbf{\underline{O}}del for \textbf{\underline{P}}ersonalized m\textbf{\underline{E}}ntal health prediction presented in this paper. 
\end{itemize}

The evaluation results are summarized in~\Cref{table_cm}. Baseline 1 achieves 60\% accuracy due to feature redundancy and lack of personalization. Baseline 2 improves to 70\% but still faces feature correlation and redundancy issues. Baseline 3 reduces feature correlation using feature importance but drops to 65\% accuracy by excluding significant features. In contrast, \SYSTEM{} reaches 91\% overall prediction accuracy by using interaction labels to capture distinct user behaviors while balancing interpretability and predictive power. We can see that the baseline methods struggle with feature redundancy and adapting to user-specific patterns, while \SYSTEM{} effectively integrates personalized interaction labels and feature importance. We also breakdown \SYSTEM{}'s predictions against the true labels for each category using a confusion matrix. As shown in \Cref{fig_cm}, \SYSTEM{} achieves high accuracy across all four categories. Overall, the evaluation using multiple metrics highlights the strong predictive performance of \SYSTEM{}.

\subsection{Interpretability}

To gain insights into mental health and improve prediction accuracy, it is crucial to extract meaningful and interpretable features from raw behavioral data. \SYSTEM{} simplifies and organizes raw data into actionable constructs, enabling the capture of individual behavioral patterns and their connection to mental health outcomes.

\begin{figure*}
    \centering
    \includegraphics[width=1\linewidth]{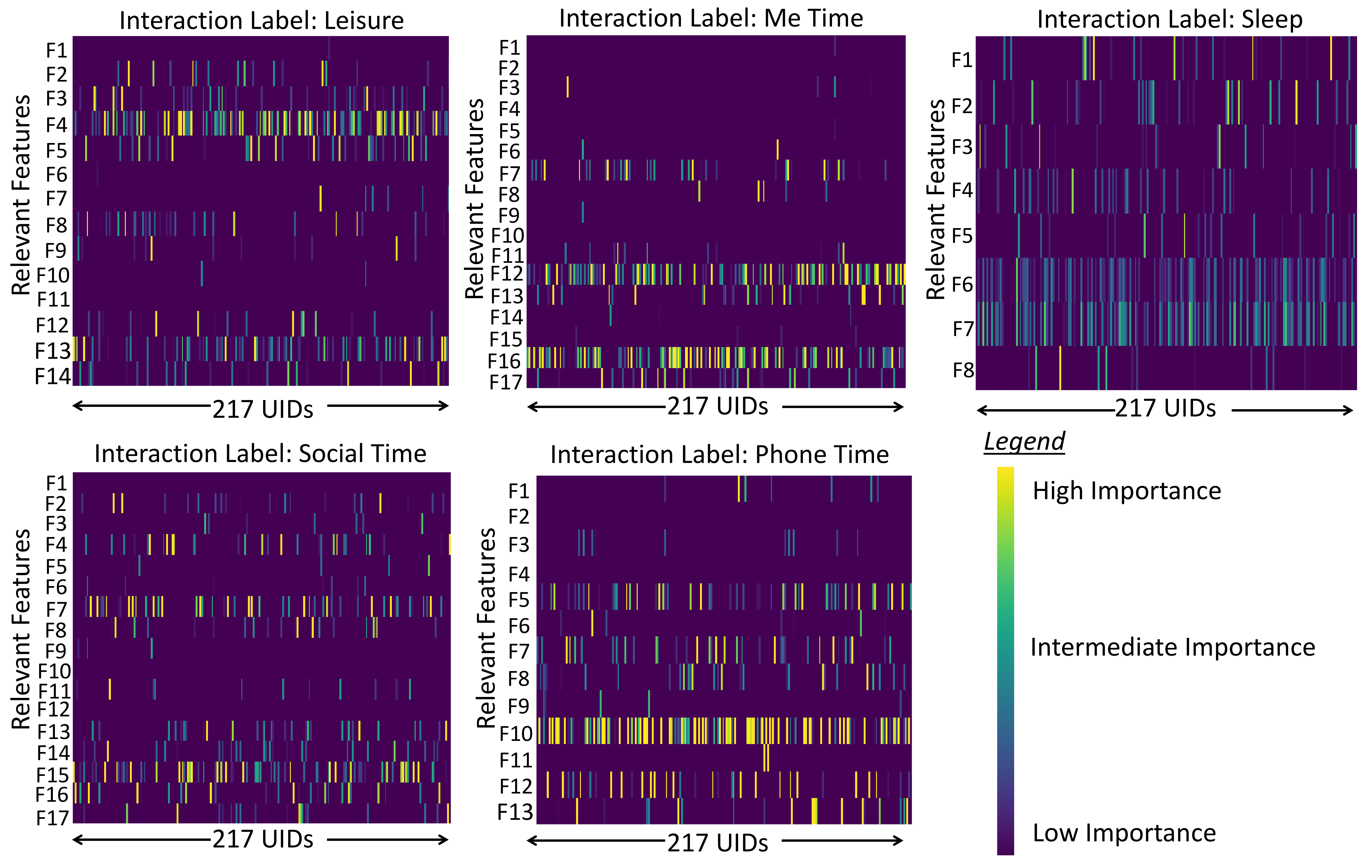}
    \caption{\textbf{The individual-level heatmaps that highlight feature importance for each interaction label across 217 individuals (UIDs).} The y-axis represents label-specific features, which vary across different interaction labels.}
    \label{fig_featureimp1}
\end{figure*}

\subsubsection{Evaluation of Stage 1: Input Features $\rightarrow$ Interaction Labels}

The first stage of \SYSTEM{} maps the raw input features into five interaction labels: \emph{Leisure}, \emph{Me Time}, \emph{Phone Time}, \emph{Sleep}, and \emph{Social Time}. This simplifies the dataset while preserving key behavioral patterns relevant to mental health. To enhance interpretability, we conduct feature importance analysis for all selected features within each label. This analysis identifies which features most influence the model's predictions, revealing meaningful behavioral patterns related to mental health. We use a random forest model to compute feature importance scores, providing an interpretable ranking of features based on their contributions. This analysis is performed individually for personalized insights into user behavior.

\Cref{fig_featureimp1} shows personalized feature importance across five labels using heatmaps for 217 students. The analysis reveals distinct patterns of feature relevance for each label, highlighting both shared and unique predictors. Below, we summarize the key findings and insights for each label.

\begin{itemize}
    \item \textbf{Label: Leisure -} Feature F4 (Duration of walking) emerged as the most important predictor for 90.36\% of individuals, emphasizing walking as a key leisure activity. Other features, such as F13 (Phone usage at home), F3 (Duration of running), F5 (Duration of conversations), F8 (Voice detection at others' dorms), and F14 (Phone usage at others' dorms), are moderately important for 20–25\% of individuals, indicating that leisure also encompasses social interactions and phone use. In contrast, features like F1 (Biking duration), F2 (Footsteps), F6 (Duration of calls), F9 (Conversations at others' dorms), F10 (Audio detections at social places), F11 (Duration at ``leisure''), and F12 (Workout duration) consistently show lower relevance. F11 has low importance despite its direct link to the label, likely due to unclear dataset definitions, underscoring the need for improved data clarity. \emph{These findings illustrate how behaviors like walking and social interactions influence leisure and provide insights into broader themes related to mental health and lifestyle habits.   }
    
    \item \textbf{Label: Me Time -} Features such as F16 (Phone usage at own dorm), F12 (Time spent at own dorm), F7 (Voice detections at own dorm), F13 (Study duration), and F17 (Phone usage at study locations) are important for 89.5\% of individuals, with F12 and F16 being the most significant. These findings suggest that activities in dorm spaces, such as studying and socializing, are key indicators of ``Me Time'' and contribute to a supportive environment. Conversely, features like F1 (Biking duration), F2 (Footsteps), F3 (Running duration), F5 (Walking duration), and F14 (Workout duration) show low importance, indicating that individual physical activities are less relevant to this label. Other features display varying levels of significance across individuals, reflecting the diverse experiences of ``Me Time''. \emph{These results underscore the role of personal environments like dorms in fostering meaningful routines and behaviors, offering insights into how individuals balance solitude, productivity, and emotional wellbeing.}
 
    \item \textbf{Label: Sleep - } Features F6 (Sleep duration) and F7 (Duration of being idle) are key predictors, highlighting the importance of inactivity and sleep in identifying sleep patterns. Features like F1 (Conversation detection at dorm) and F4 (Location at dorm) show moderate importance, suggesting that reduced social interactions and staying in the dorm contribute to recognizing sleep behaviors, likely linked to quiet nighttime routines. In contrast, features such as F2 (Voice detections at home), F3 (Conversations at home), F5 (Phone usage at home), and F8 (Health fitness) are minimally relevant. \emph{These results emphasize the role of inactivity and dorm contexts in understanding sleep patterns while activities like fitness or phone use have a limited impact on sleep.} 

    \item \textbf{Label: Social Time - }  Feature F15 (Duration at study location) is the key predictor, indicating a strong link between study locations and social activities. Other features include F2 (Walking duration), F7 (Voice detections at study areas), F13 (Time spent in others' dorms), and F16 (Phone usage in others' dorms), which highlight the importance of mobility, conversations, and phone use. In contrast, features like F1 (Footsteps), F9 (Conversations in others' dorms), F10 (Conversations at social places), and F12 (Duration of leisure) show limited importance, suggesting they are less directly related to social time or relevant only for a few individuals. \emph{These findings emphasize the role of specific environments and interaction patterns in shaping social time while indicating that factors like leisure duration or general mobility have a minor impact.}   

    \item \textbf{Label: Phone Time - } Feature F10 (Phone usage at own dorm) is the primary predictor, indicating that students primarily use their phones in personal spaces where they feel comfortable engaging in calls and messaging. Other features include F5 (Audio detected at own dorm), F7 (Audio detected at home), F8 (Phone usage at home), F12 (Phone usage at study locations), and F13 (Phone usage throughout the day), which highlight variations in phone use across personal, home, and study contexts. Conversely, features like F1 (Audio conversations detected throughout the day), F2 (Number of SMS), F3 (Voice detection at home), F4 (Voice detection in social settings), F6 (Conversations detected at home), F9 (Phone usage in others' dorms), and F10 (Phone usage in workout areas) show minimal relevance, likely due to ambiguous definitions or differences in data collection between iOS and Android devices. \emph{These findings emphasize the role of personal and study spaces in shaping phone usage while highlighting the need for clearer feature definitions and standardized data collection to enhance the reliability of less relevant features.}
\end{itemize}

In summary, the findings reveal personalized variability among students in feature importance, with key behavioral patterns linked to mental health and activity contexts:

\begin{itemize}
    \item \textbf{Behavioral patterns:}  
    Walking, phone usage, and time in personal spaces are key to labels like \emph{Leisure}, \emph{Phone Time}, and \emph{Me Time}. Walking supports relaxation, while dorm-based phone use reflects connection and self-reflection, linking physical spaces to emotional wellbeing.
    \item \textbf{Activity contexts:}  Study locations and others' dorms strongly predict \emph{Social Time}, emphasizing collaborative settings. In contrast, private spaces dominate \emph{Me Time} and \emph{Phone Time}, illustrating the balance between shared and personal spaces for social engagement and self-care.
    \item \textbf{Sleep and rest:} 
    Sleep duration and idle time are central to \emph{Sleep}, with dorm-related contexts suggesting restful routines tied to spatial habits and reduced interactions, reinforcing sleep's role in mental health.
    \item \textbf{Personalized variability as a key factor:}  
    The varying importance of features like walking or conversation detection reflects the individual nature of behavior. Tailored analyses can enhance mental health predictions, making them more accurate and context-aware.
\end{itemize}

These results emphasize the importance of selecting features that capture distinct behavioral dimensions for interaction labels. They also demonstrate that personalized analysis is key for accurate mental health prediction, as feature importance varies across users. This forms the basis for our label scoring and predictions, explained in the following subsections.

\begin{figure}[t]
    \centering
    \includegraphics[width=1\linewidth]{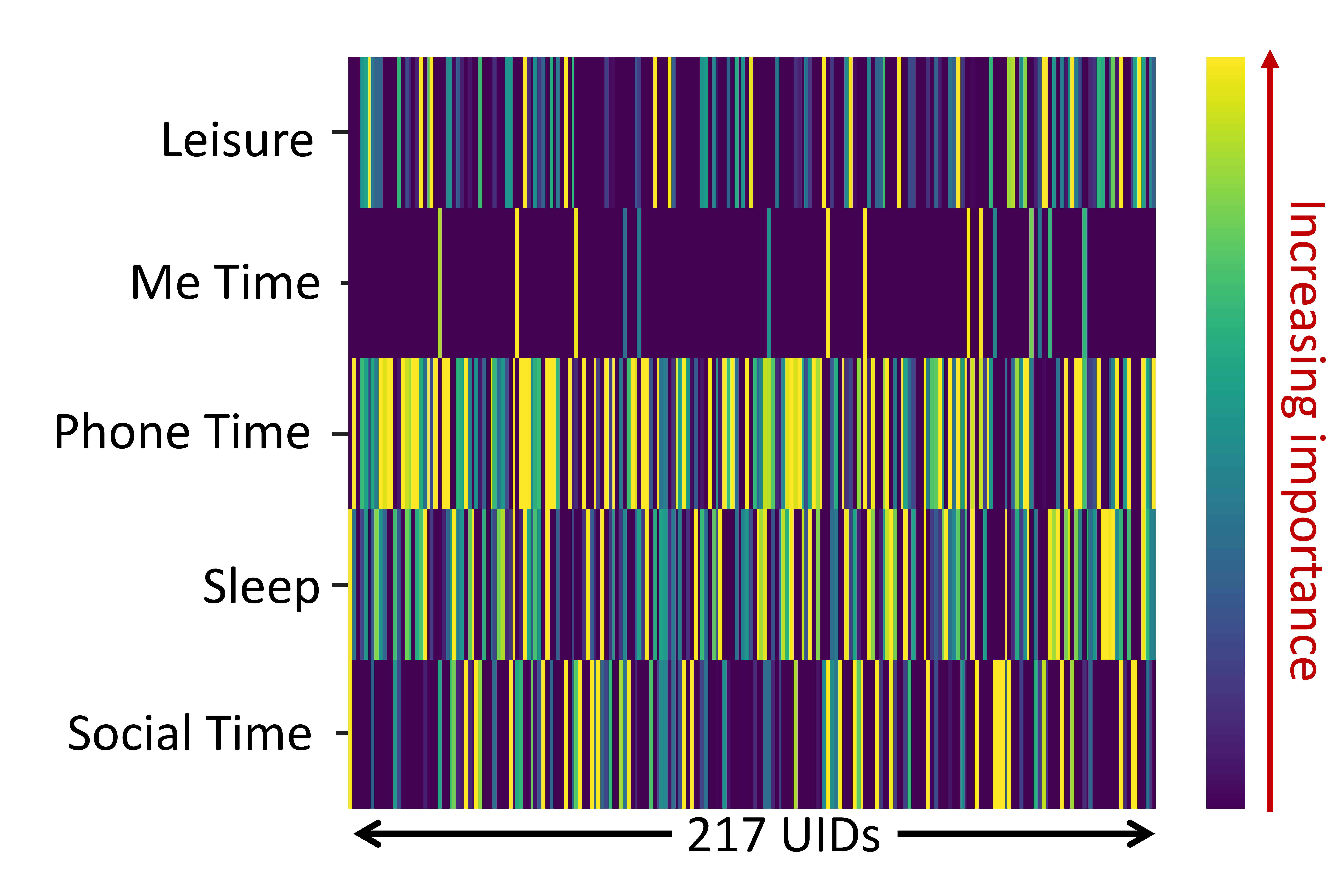}
    \caption{\textbf{The individual-level heatmap that shows the importance of interaction label scores in predicting PHQ-4 categories.} Each row represents an interaction label across 217 individuals.}
    \label{fig_featureimp_all}
\end{figure}

\subsubsection{Evaluation of Stage 2: Interaction Labels $\rightarrow$ PHQ-4 Categories}

\Cref{fig_featureimp_all} shows the importance of interaction labels (\emph{Leisure, Me Time, Phone Time, Sleep, Social Time}) in predicting PHQ-4 mental health categories across 217 users. Each row corresponds to a label, and each column represents an individual user, with colors ranging from dark purple (low importance) to bright yellow (high importance). The figure highlights how the relevance of these labels varies significantly across individuals, revealing the personalized nature of mental health predictors.

Among the labels, \emph{Sleep} is the most consistently important, with high relevance for 95\% of students, reflecting the strong link between healthy sleep patterns and stable mental health. \emph{Phone Time} also shows significant importance across many users, likely indicating stress or coping behaviors related to screen use. In contrast, labels like \emph{Social Time} and \emph{Leisure} exhibit more variability; \emph{Social Time} is crucial for some but less so for others, reflecting individual differences in social interactions' impact on mental health. Similarly, \emph{Leisure} is an important predictor for a subset of users, emphasizing the role of relaxation activities. However, \emph{Me Time} serves as a predictor for an even smaller group, highlighting the challenges of finding solitude in a school environment with shared spaces.

In all, \Cref{fig_featureimp_all} underscores a critical insight -- mental health predictors are not one-size-fits-all. Each label contributes uniquely, reflecting the diverse and personalized ways behavioral patterns influence mental health outcomes. While \emph{Sleep} and \emph{Phone Time} act as general predictors across most users, \emph{Social Time}, \emph{Me Time}, and \emph{Leisure} provide insights tailored to specific individuals.
By combining these labels, we can build a comprehensive understanding of behavioral patterns and their impact on mental health, supporting the need for personalized modeling approaches to predict PHQ-4 categories effectively.
\begin{figure}[t]
    \centering
    \includegraphics[width=1\linewidth]{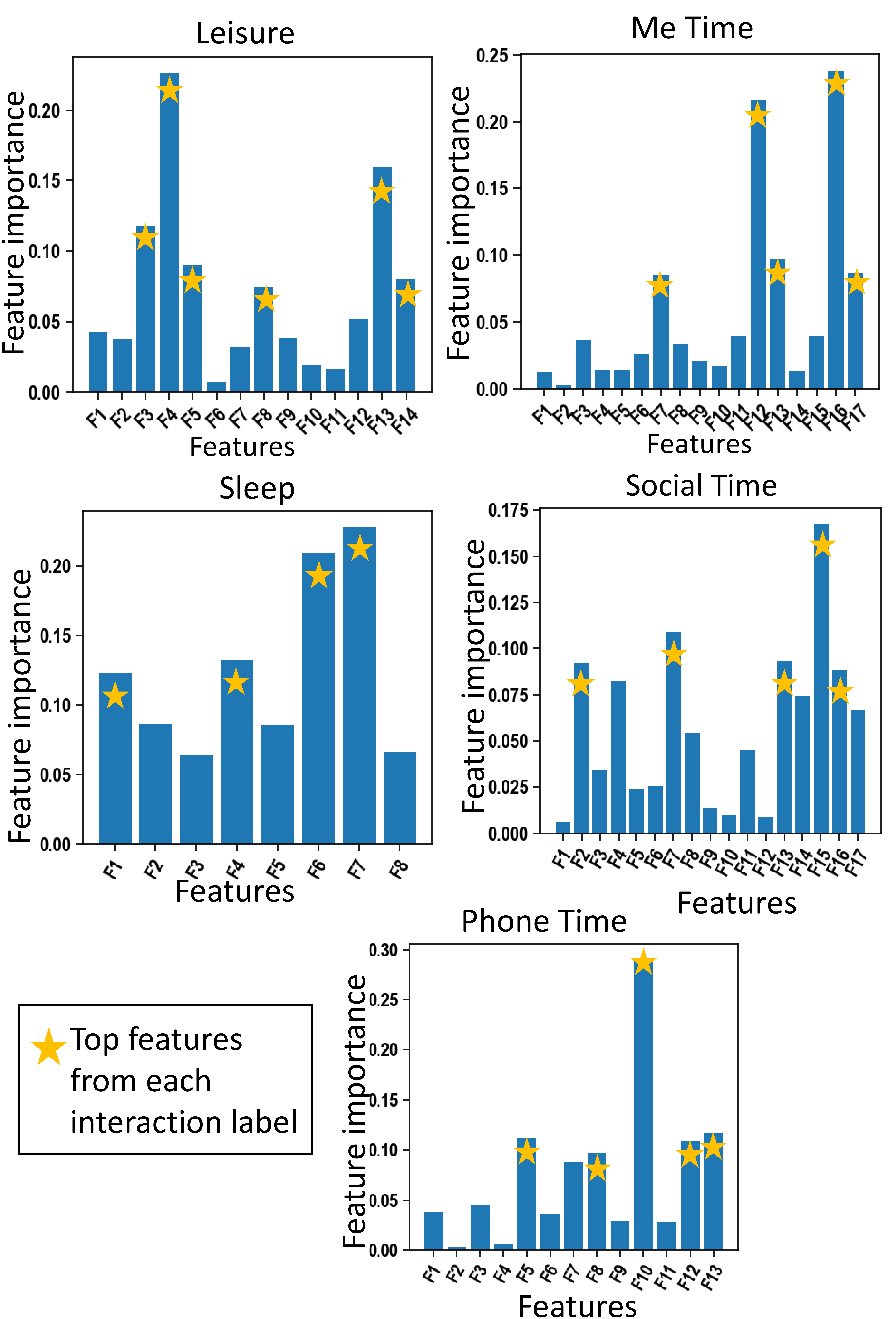}
    \caption{\textbf{Mean feature importance across all individuals for each interaction label.} }
    \label{fig_featureimp_mean}
\end{figure}

\subsection{Scoring for Interaction Labels}

To analyze how each feature contributes to scoring for each interaction label, we visualize the feature importances of relevant features derived from random forest models across all individuals. \Cref{fig_featureimp_mean} displays the top features (marked with a star) for each interaction label:
\begin{itemize}
    \item \textbf{Leisure:} F3 (running duration), F4 (walking duration) , F5 (conversation duration), F8 (voice detection at others' dorms), F13 (phone usage at home), and F14 (phone usage at others' dorms).
    \item \textbf{Me Time:} F7 (voice detections at own dorm), F12 (time spent in own dorm), F13 (study duration), F16 (phone usage at own dorm), and F17 (phone usage at study locations).
    \item \textbf{Sleep:} F1 (conversations detected at dorm), F4 (duration at own dorm), F6 (sleep duration), and F7 (duration of being idle).
    \item \textbf{Social Time:} F2 (walking duration), F7 (voice detected at study location), F13 (time in others' dorms), F15 (duration at study location), and F16 (phone usage at others' dorm).
    \item \textbf{Phone Time:} F5 (audio detected at own dorm), F8 (phone usage at home), F10 (phone usage at own dorm), F12 (phone usage at study locations), and F13 (phone usage throughout the day). Phone usage corresponds to the ratio of the number of phone unlocks to usage duration. 
\end{itemize}

As illustrated in~\Cref{alg:sleep-final}, the scores for each interaction label are calculated based on the weighted importance of various features (i.e., NWFI in \Cref{alg:sleep-final}), measured against a threshold value derived from the mean of distribution across all individuals. Next, we show the thresholds for \emph{Leisure Time} and \emph{Sleep} to demonstrate how feature contributions influence the final label scoring.

\begin{figure}[t]
    \centering
    \includegraphics[width=1\linewidth]
    {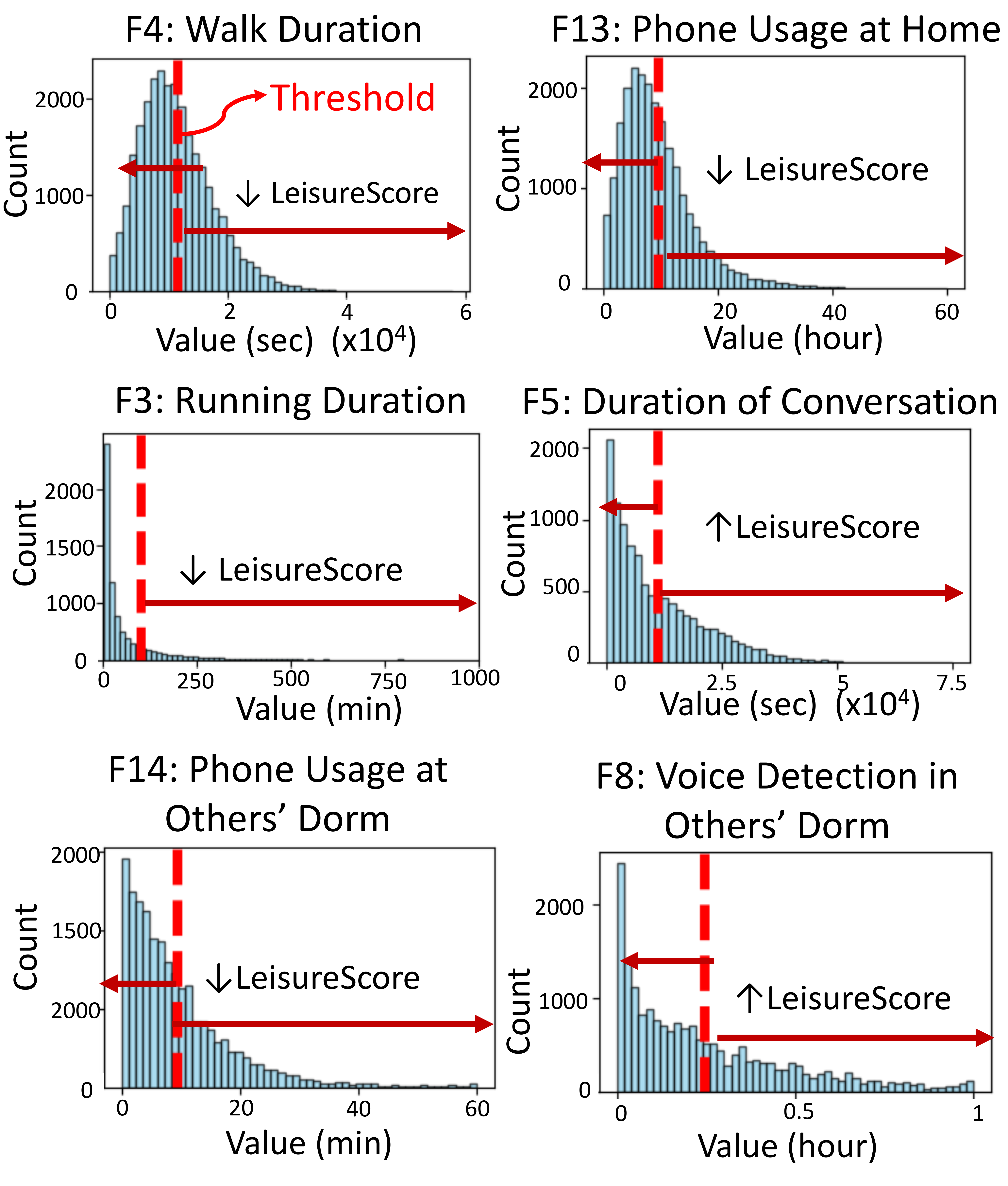}
    \caption{\textbf{Thresholds for LeisureScore calculation based on six features.} The vertical dashed red line is the threshold.}
    \label{fig_leisurethreshold}
\end{figure}

\begin{figure}[t]
    \centering
    \includegraphics[width=1\linewidth]{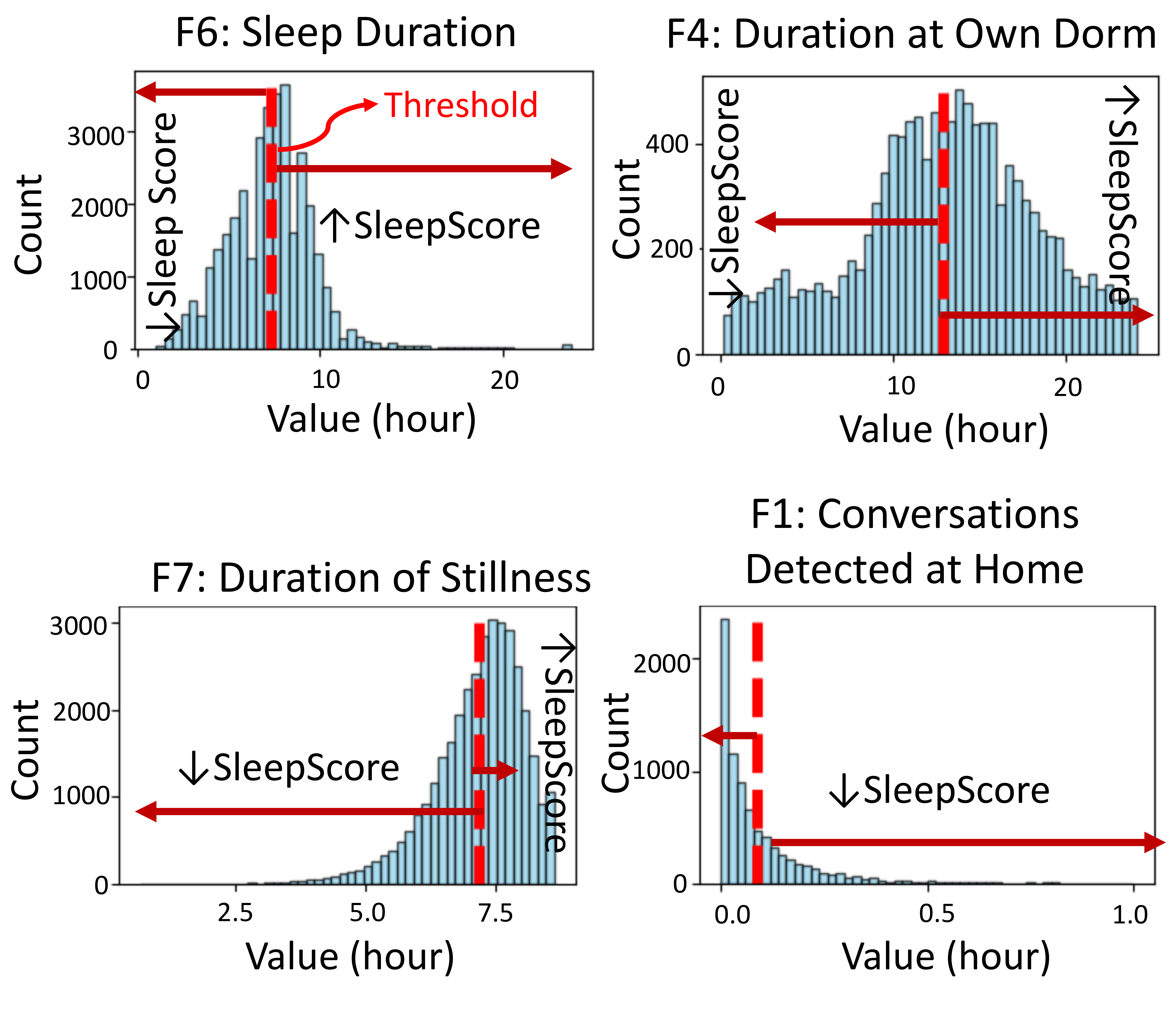}
    \caption{\textbf{Thresholds for SleepScore calculation based on four features.} The vertical dashed red line is the threshold.}
    \label{fig_sleepthreshold}
\end{figure}

In \Cref{fig_leisurethreshold}, the dotted lines highlight the threshold values for the most important features (F4, F13, F3, F5, F14, and F8) for \emph{Leisure Time} . F4 reflects walking duration: if a student's walking duration is less than the threshold (3 hours), it indicates leisure activity. Consequently, the \emph{LeisureScore} is incremented proportionally to the weighted feature importance of walking for the respective student. Similarly, F8 considers voice detection in others' dormitories. If the duration exceeds the threshold (15 minutes), it suggests that the student is having leisure time with friends, causing an increase in \emph{LeisureScore} by the corresponding weighted feature importance. 

In \Cref{fig_sleepthreshold}, the dotted lines highlight the threshold values for the most important features (F6, F4, F7, and F1) for \emph{Sleep}. F6 indicates total sleep duration: if a student's sleep exceeds 7 hours, representing the standard benchmark for healthy sleep, the \emph{SleepScore} is incremented based on the student's weighted feature importance for sleep. F4, which measures the time spent in the room, shows that durations greater than 12.9 hours typically signify overnight dorm occupancy, often indicative of sleep rather than other activities. F7 corroborates healthy sleep with thresholds above 7.5 hours, while F1 reflects quietness, with values below 6 minutes suggesting an environment conducive to rest. 

%% file: conclusion.tex
\section{Related Work}

Machine learning increasingly tackles various challenges in mental health research~\cite{shatte2019machine,d2020ai,kim2021machine,sajno2023machine,tutun2023ai,adler2022machine}. Using data from mobile devices that monitor smartphone use, social media activity, and medical records, researchers have developed machine learning techniques to identify and predict conditions such as depression, anxiety, and stress~\cite{saeb2015mobile,guntuku2017detecting,zhao2024digital}. Furthermore, natural language processing has opened new pathways for analyzing text data, including social media posts and therapy transcripts, offering valuable insights into emotional and psychological states~\cite{eichstaedt2018facebook,althoff2016large,cook2016novel,stewart2021applied,lamichhane2023evaluation}.

However, despite these advances, many machine learning techniques struggle to capture and understand mental health characteristics at the individual level. Researchers often focus on applying techniques without addressing the issues of variability, personalization, and interpretability~\cite{mohr2017personal,torous2016new,batstra2024more,moggia2024treatment}. This limitation restricts the practical application of these methods, particularly in treatment recommendations and interventions. 

The release of the College Experience Study dataset offers a unique opportunity to analyze the longest longitudinal passive mobile sensing dataset on the mental health of college students at the individual level~\cite{nepal2024capturing}. Our prior work was the first attempt to analyze this dataset~\cite{xuan2025unlocking}. This paper goes a step further by addressing a key research gap with \SYSTEM{}, the first interpretable hierarchical model designed to accurately predict and offer new insights into the mental health of college students.


\section{Conclusion and Future Work}

This paper addresses the challenge of personalized predictions and understanding mental health among college students. Leveraging a cutting-edge longitudinal dataset, we present \SYSTEM{}, the first hierarchical model designed to predict and interpret mental health status at the individual level. \SYSTEM{} bridges the gap between black-box machine learning and actionable insights, offering both accurate and interpretable, personalized predictions. These contributions are especially important in the mental health domain, where  individualization is critical.

Our evaluation shows that \SYSTEM{} has a better performance than baseline methods and offers insights into how behavioral patterns, such as sleep, phone usage, and social time, relate to mental health outcomes on a personalized level. However, like any data-driven model, there are several areas that offer opportunities for further improvement, which we describe as follows. 

From a data perspective, our analysis relied on features aggregated at a daily level. Incorporating higher-resolution hourly data could reveal more granular behavioral patterns, potentially enhancing prediction accuracy. Additionally, certain data processing decisions—such as replacing missing values with zeros—may oversimplify real-world behaviors. Future work could explore more sophisticated imputation techniques to better capture underlying trends.

From a modeling perspective, while the current framework uses five interaction labels—empirically motivated and preliminarily validated—future work could systematically explore how varying the number of labels impacts both predictive performance and interpretability. Additionally, incorporating aggregate metrics (e.g., mean performance) and confidence intervals during evaluation would offer a more robust assessment of model reliability.  

Furthermore, the lack of a fixed random seed introduced unintended variability in reported results; standardizing this in future experiments will improve reproducibility and consistency.  

These refinements present a pathway for advancing mental health research, enabling the development of more interpretable, personalized, and actionable models.

\begin{acks}

We thank anonymous reviewers for their helpful feedback to improve the final version of the paper. The work of Meghna Roy Chowdhury and Shreyas Sen is supported by the National Science Foundation under Grant \# 1944602.

\end{acks}

%% file: main.bbl

\begin{thebibliography}{40}


\ifx \showCODEN    \undefined \def \showCODEN     #1{\unskip}     \fi
\ifx \showISBNx    \undefined \def \showISBNx     #1{\unskip}     \fi
\ifx \showISBNxiii \undefined \def \showISBNxiii  #1{\unskip}     \fi
\ifx \showISSN     \undefined \def \showISSN      #1{\unskip}     \fi
\ifx \showLCCN     \undefined \def \showLCCN      #1{\unskip}     \fi
\ifx \shownote     \undefined \def \shownote      #1{#1}          \fi
\ifx \showarticletitle \undefined \def \showarticletitle #1{#1}   \fi
\ifx \showURL      \undefined \def \showURL       {\relax}        \fi
\providecommand\bibfield[2]{#2}
\providecommand\bibinfo[2]{#2}
\providecommand\natexlab[1]{#1}
\providecommand\showeprint[2][]{arXiv:#2}

\bibitem[Abrams(2022)]%
        {abrams2022student}
\bibfield{author}{\bibinfo{person}{Zara Abrams}.} \bibinfo{year}{2022}\natexlab{}.
\newblock \showarticletitle{Student mental health is in crisis. Campuses are rethinking their approach}.
\newblock \bibinfo{journal}{\emph{Monitor on Psychology}} \bibinfo{volume}{53}, \bibinfo{number}{7} (\bibinfo{year}{2022}), \bibinfo{pages}{60}.
\newblock


\bibitem[Adler et~al\mbox{.}(2022)]%
        {adler2022machine}
\bibfield{author}{\bibinfo{person}{Daniel~A Adler}, \bibinfo{person}{Fei Wang}, \bibinfo{person}{David~C Mohr}, {and} \bibinfo{person}{Tanzeem Choudhury}.} \bibinfo{year}{2022}\natexlab{}.
\newblock \showarticletitle{Machine learning for passive mental health symptom prediction: Generalization across different longitudinal mobile sensing studies}.
\newblock \bibinfo{journal}{\emph{Plos one}} \bibinfo{volume}{17}, \bibinfo{number}{4} (\bibinfo{year}{2022}), \bibinfo{pages}{e0266516}.
\newblock


\bibitem[Althoff et~al\mbox{.}(2016)]%
        {althoff2016large}
\bibfield{author}{\bibinfo{person}{Tim Althoff}, \bibinfo{person}{Kevin Clark}, {and} \bibinfo{person}{Jure Leskovec}.} \bibinfo{year}{2016}\natexlab{}.
\newblock \showarticletitle{Large-scale analysis of counseling conversations: An application of natural language processing to mental health}.
\newblock \bibinfo{journal}{\emph{Transactions of the Association for Computational Linguistics}}  \bibinfo{volume}{4} (\bibinfo{year}{2016}), \bibinfo{pages}{463--476}.
\newblock


\bibitem[Barbayannis et~al\mbox{.}(2022)]%
        {barbayannis2022academic}
\bibfield{author}{\bibinfo{person}{Georgia Barbayannis}, \bibinfo{person}{Mahindra Bandari}, \bibinfo{person}{Xiang Zheng}, \bibinfo{person}{Humberto Baquerizo}, \bibinfo{person}{Keith~W Pecor}, {and} \bibinfo{person}{Xue Ming}.} \bibinfo{year}{2022}\natexlab{}.
\newblock \showarticletitle{Academic stress and mental well-being in college students: correlations, affected groups, and COVID-19}.
\newblock \bibinfo{journal}{\emph{Frontiers in psychology}}  \bibinfo{volume}{13} (\bibinfo{year}{2022}), \bibinfo{pages}{886344}.
\newblock


\bibitem[Batstra and Timimi(2024)]%
        {batstra2024more}
\bibfield{author}{\bibinfo{person}{Laura Batstra} {and} \bibinfo{person}{Sami Timimi}.} \bibinfo{year}{2024}\natexlab{}.
\newblock \showarticletitle{Is more psychotherapy a dead horse? An essay on the (in) effectiveness of individual treatment for mental suffering}.
\newblock \bibinfo{journal}{\emph{PLOS Mental Health}} \bibinfo{volume}{1}, \bibinfo{number}{7} (\bibinfo{year}{2024}), \bibinfo{pages}{e0000194}.
\newblock


\bibitem[BestColleges(2024)]%
        {bestcolleges2024}
\bibfield{author}{\bibinfo{person}{BestColleges}.} \bibinfo{year}{2024}\natexlab{}.
\newblock \bibinfo{title}{College Student Mental Health Statistics}.
\newblock \bibinfo{howpublished}{\url{https://www.bestcolleges.com/research/college-student-mental-health-statistics/}}.
\newblock
\newblock
\shownote{Accessed: 2024-03-22}.


\bibitem[Breiman(2001)]%
        {breiman2001random}
\bibfield{author}{\bibinfo{person}{Leo Breiman}.} \bibinfo{year}{2001}\natexlab{}.
\newblock \showarticletitle{Random forests}.
\newblock \bibinfo{journal}{\emph{Machine learning}}  \bibinfo{volume}{45} (\bibinfo{year}{2001}), \bibinfo{pages}{5--32}.
\newblock


\bibitem[Chu et~al\mbox{.}(2023)]%
        {chu2023association}
\bibfield{author}{\bibinfo{person}{Tianshu Chu}, \bibinfo{person}{Xin Liu}, \bibinfo{person}{Shigemi Takayanagi}, \bibinfo{person}{Tomoko Matsushita}, {and} \bibinfo{person}{Hiro Kishimoto}.} \bibinfo{year}{2023}\natexlab{}.
\newblock \showarticletitle{Association between mental health and academic performance among university undergraduates: The interacting role of lifestyle behaviors}.
\newblock \bibinfo{journal}{\emph{International Journal of Methods in Psychiatric Research}} \bibinfo{volume}{32}, \bibinfo{number}{1} (\bibinfo{year}{2023}), \bibinfo{pages}{e1938}.
\newblock


\bibitem[Colarossi(2022)]%
        {colarossi2022mental}
\bibfield{author}{\bibinfo{person}{Jessica Colarossi}.} \bibinfo{year}{2022}\natexlab{}.
\newblock \showarticletitle{Mental health of college students is getting worse}.
\newblock \bibinfo{journal}{\emph{The Brink}} (\bibinfo{year}{2022}).
\newblock


\bibitem[Conley(2024)]%
        {Findingp69:online}
\bibfield{author}{\bibinfo{person}{Mark Conley}.} \bibinfo{year}{2024}\natexlab{}.
\newblock \bibinfo{title}{Finding personalized approaches to treating mental illness}.
\newblock
\urldef\tempurl%
\url{https://stanmed.stanford.edu/precision-mental-health-promise/}
\showURL{%
\tempurl}
\newblock
\shownote{[Online; accessed 2025-01-04]}.


\bibitem[Cook et~al\mbox{.}(2016)]%
        {cook2016novel}
\bibfield{author}{\bibinfo{person}{Benjamin~L Cook}, \bibinfo{person}{Ana~M Progovac}, \bibinfo{person}{Pei Chen}, \bibinfo{person}{Brian Mullin}, \bibinfo{person}{Sherry Hou}, {and} \bibinfo{person}{Enrique Baca-Garcia}.} \bibinfo{year}{2016}\natexlab{}.
\newblock \showarticletitle{Novel use of natural language processing (NLP) to predict suicidal ideation and psychiatric symptoms in a text-based mental health intervention in Madrid}.
\newblock \bibinfo{journal}{\emph{Computational and mathematical methods in medicine}} \bibinfo{volume}{2016}, \bibinfo{number}{1} (\bibinfo{year}{2016}), \bibinfo{pages}{8708434}.
\newblock


\bibitem[D’Alfonso(2020)]%
        {d2020ai}
\bibfield{author}{\bibinfo{person}{Simon D’Alfonso}.} \bibinfo{year}{2020}\natexlab{}.
\newblock \showarticletitle{AI in mental health}.
\newblock \bibinfo{journal}{\emph{Current opinion in psychology}}  \bibinfo{volume}{36} (\bibinfo{year}{2020}), \bibinfo{pages}{112--117}.
\newblock


\bibitem[Eichstaedt et~al\mbox{.}(2018)]%
        {eichstaedt2018facebook}
\bibfield{author}{\bibinfo{person}{Johannes~C Eichstaedt}, \bibinfo{person}{Robert~J Smith}, \bibinfo{person}{Raina~M Merchant}, \bibinfo{person}{Lyle~H Ungar}, \bibinfo{person}{Patrick Crutchley}, \bibinfo{person}{Daniel Preo{\c{t}}iuc-Pietro}, \bibinfo{person}{David~A Asch}, {and} \bibinfo{person}{H~Andrew Schwartz}.} \bibinfo{year}{2018}\natexlab{}.
\newblock \showarticletitle{Facebook language predicts depression in medical records}.
\newblock \bibinfo{journal}{\emph{Proceedings of the National Academy of Sciences}} \bibinfo{volume}{115}, \bibinfo{number}{44} (\bibinfo{year}{2018}), \bibinfo{pages}{11203--11208}.
\newblock


\bibitem[Elamin et~al\mbox{.}(2024)]%
        {elamin2024smartphone}
\bibfield{author}{\bibinfo{person}{Nadia~O Elamin}, \bibinfo{person}{Juman~M Almasaad}, \bibinfo{person}{Reem~B Busaeed}, \bibinfo{person}{Daniah~A Aljafari}, {and} \bibinfo{person}{Muhammed~A Khan}.} \bibinfo{year}{2024}\natexlab{}.
\newblock \showarticletitle{Smartphone addiction, stress, and depression among university students}.
\newblock \bibinfo{journal}{\emph{Clinical Epidemiology and Global Health}}  \bibinfo{volume}{25} (\bibinfo{year}{2024}), \bibinfo{pages}{101487}.
\newblock


\bibitem[Guntuku et~al\mbox{.}(2017)]%
        {guntuku2017detecting}
\bibfield{author}{\bibinfo{person}{Sharath~Chandra Guntuku}, \bibinfo{person}{David~B Yaden}, \bibinfo{person}{Margaret~L Kern}, \bibinfo{person}{Lyle~H Ungar}, {and} \bibinfo{person}{Johannes~C Eichstaedt}.} \bibinfo{year}{2017}\natexlab{}.
\newblock \showarticletitle{Detecting depression and mental illness on social media: an integrative review}.
\newblock \bibinfo{journal}{\emph{Current Opinion in Behavioral Sciences}}  \bibinfo{volume}{18} (\bibinfo{year}{2017}), \bibinfo{pages}{43--49}.
\newblock


\bibitem[Hammoudi~Halat et~al\mbox{.}(2023)]%
        {hammoudi2023understanding}
\bibfield{author}{\bibinfo{person}{Dalal Hammoudi~Halat}, \bibinfo{person}{Abderrezzaq Soltani}, \bibinfo{person}{Roua Dalli}, \bibinfo{person}{Lama Alsarraj}, {and} \bibinfo{person}{Ahmed Malki}.} \bibinfo{year}{2023}\natexlab{}.
\newblock \showarticletitle{Understanding and fostering mental health and well-being among university faculty: A narrative review}.
\newblock \bibinfo{journal}{\emph{Journal of clinical medicine}} \bibinfo{volume}{12}, \bibinfo{number}{13} (\bibinfo{year}{2023}), \bibinfo{pages}{4425}.
\newblock


\bibitem[He et~al\mbox{.}(2023)]%
        {he2023personalized}
\bibfield{author}{\bibinfo{person}{Zengyou He}, \bibinfo{person}{Yifan Tang}, \bibinfo{person}{Lianyu Hu}, \bibinfo{person}{Mudi Jiang}, {and} \bibinfo{person}{Yan Liu}.} \bibinfo{year}{2023}\natexlab{}.
\newblock \showarticletitle{Personalized Interpretable Classification}.
\newblock \bibinfo{journal}{\emph{arXiv preprint arXiv:2302.02528}} (\bibinfo{year}{2023}).
\newblock


\bibitem[Kim et~al\mbox{.}(2021)]%
        {kim2021machine}
\bibfield{author}{\bibinfo{person}{Jina Kim}, \bibinfo{person}{Daeun Lee}, \bibinfo{person}{Eunil Park}, {et~al\mbox{.}}} \bibinfo{year}{2021}\natexlab{}.
\newblock \showarticletitle{Machine learning for mental health in social media: bibliometric study}.
\newblock \bibinfo{journal}{\emph{Journal of Medical Internet Research}} \bibinfo{volume}{23}, \bibinfo{number}{3} (\bibinfo{year}{2021}), \bibinfo{pages}{e24870}.
\newblock


\bibitem[Kingma and Ba(2014)]%
        {kingma2014adam}
\bibfield{author}{\bibinfo{person}{Diederik~P Kingma} {and} \bibinfo{person}{Jimmy Ba}.} \bibinfo{year}{2014}\natexlab{}.
\newblock \showarticletitle{Adam: A method for stochastic optimization}.
\newblock \bibinfo{journal}{\emph{arXiv preprint arXiv:1412.6980}} (\bibinfo{year}{2014}).
\newblock


\bibitem[Kroenke et~al\mbox{.}(2009)]%
        {kroenke2009ultra}
\bibfield{author}{\bibinfo{person}{Kurt Kroenke}, \bibinfo{person}{Robert~L Spitzer}, \bibinfo{person}{Janet~BW Williams}, {and} \bibinfo{person}{Bernd L{\"o}we}.} \bibinfo{year}{2009}\natexlab{}.
\newblock \showarticletitle{An ultra-brief screening scale for anxiety and depression: the PHQ--4}.
\newblock \bibinfo{journal}{\emph{Psychosomatics}} \bibinfo{volume}{50}, \bibinfo{number}{6} (\bibinfo{year}{2009}), \bibinfo{pages}{613--621}.
\newblock


\bibitem[Lamichhane(2023)]%
        {lamichhane2023evaluation}
\bibfield{author}{\bibinfo{person}{Bishal Lamichhane}.} \bibinfo{year}{2023}\natexlab{}.
\newblock \showarticletitle{Evaluation of chatgpt for nlp-based mental health applications}.
\newblock \bibinfo{journal}{\emph{arXiv preprint arXiv:2303.15727}} (\bibinfo{year}{2023}).
\newblock


\bibitem[Lengerich et~al\mbox{.}(2019)]%
        {lengerich2019learning}
\bibfield{author}{\bibinfo{person}{Ben Lengerich}, \bibinfo{person}{Bryon Aragam}, {and} \bibinfo{person}{Eric~P Xing}.} \bibinfo{year}{2019}\natexlab{}.
\newblock \showarticletitle{Learning sample-specific models with low-rank personalized regression}.
\newblock \bibinfo{journal}{\emph{Advances in Neural Information Processing Systems}}  \bibinfo{volume}{32} (\bibinfo{year}{2019}).
\newblock


\bibitem[L{\"o}we et~al\mbox{.}(2010)]%
        {lowe20104}
\bibfield{author}{\bibinfo{person}{Bernd L{\"o}we}, \bibinfo{person}{Inka Wahl}, \bibinfo{person}{Matthias Rose}, \bibinfo{person}{Carsten Spitzer}, \bibinfo{person}{Heide Glaesmer}, \bibinfo{person}{Katja Wingenfeld}, \bibinfo{person}{Antonius Schneider}, {and} \bibinfo{person}{Elmar Br{\"a}hler}.} \bibinfo{year}{2010}\natexlab{}.
\newblock \showarticletitle{A 4-item measure of depression and anxiety: validation and standardization of the Patient Health Questionnaire-4 (PHQ-4) in the general population}.
\newblock \bibinfo{journal}{\emph{Journal of affective disorders}} \bibinfo{volume}{122}, \bibinfo{number}{1-2} (\bibinfo{year}{2010}), \bibinfo{pages}{86--95}.
\newblock


\bibitem[Moggia et~al\mbox{.}(2024)]%
        {moggia2024treatment}
\bibfield{author}{\bibinfo{person}{Danilo Moggia}, \bibinfo{person}{Wolfgang Lutz}, \bibinfo{person}{Eva-Lotta Brakemeier}, {and} \bibinfo{person}{Leonard Bickman}.} \bibinfo{year}{2024}\natexlab{}.
\newblock \showarticletitle{Treatment Personalization and Precision Mental Health Care: Where are we and where do we want to go?}
\newblock \bibinfo{journal}{\emph{Administration and Policy in Mental Health and Mental Health Services Research}} \bibinfo{volume}{51}, \bibinfo{number}{5} (\bibinfo{year}{2024}), \bibinfo{pages}{611--616}.
\newblock


\bibitem[Mohr et~al\mbox{.}(2017)]%
        {mohr2017personal}
\bibfield{author}{\bibinfo{person}{David~C Mohr}, \bibinfo{person}{Mi Zhang}, {and} \bibinfo{person}{Stephen~M Schueller}.} \bibinfo{year}{2017}\natexlab{}.
\newblock \showarticletitle{Personal sensing: understanding mental health using ubiquitous sensors and machine learning}.
\newblock \bibinfo{journal}{\emph{Annual review of clinical psychology}} (\bibinfo{year}{2017}).
\newblock


\bibitem[Nepal et~al\mbox{.}(2024)]%
        {nepal2024capturing}
\bibfield{author}{\bibinfo{person}{Subigya Nepal}, \bibinfo{person}{Wenjun Liu}, \bibinfo{person}{Arvind Pillai}, \bibinfo{person}{Weichen Wang}, \bibinfo{person}{Vlado Vojdanovski}, \bibinfo{person}{Jeremy~F Huckins}, \bibinfo{person}{Courtney Rogers}, \bibinfo{person}{Meghan~L Meyer}, {and} \bibinfo{person}{Andrew~T Campbell}.} \bibinfo{year}{2024}\natexlab{}.
\newblock \showarticletitle{Capturing the College Experience: A Four-Year Mobile Sensing Study of Mental Health, Resilience and Behavior of College Students during the Pandemic}.
\newblock \bibinfo{journal}{\emph{Proceedings of the ACM on Interactive, Mobile, Wearable and Ubiquitous Technologies}} \bibinfo{volume}{8}, \bibinfo{number}{1} (\bibinfo{year}{2024}), \bibinfo{pages}{1--37}.
\newblock


\bibitem[Popescu et~al\mbox{.}(2009)]%
        {popescu2009multilayer}
\bibfield{author}{\bibinfo{person}{Marius-Constantin Popescu}, \bibinfo{person}{Valentina~E Balas}, \bibinfo{person}{Liliana Perescu-Popescu}, {and} \bibinfo{person}{Nikos Mastorakis}.} \bibinfo{year}{2009}\natexlab{}.
\newblock \showarticletitle{Multilayer perceptron and neural networks}.
\newblock \bibinfo{journal}{\emph{WSEAS Transactions on Circuits and Systems}} \bibinfo{volume}{8}, \bibinfo{number}{7} (\bibinfo{year}{2009}), \bibinfo{pages}{579--588}.
\newblock


\bibitem[Saeb et~al\mbox{.}(2015)]%
        {saeb2015mobile}
\bibfield{author}{\bibinfo{person}{Sohrab Saeb}, \bibinfo{person}{Mi Zhang}, \bibinfo{person}{Christopher~J Karr}, \bibinfo{person}{Stephen~M Schueller}, \bibinfo{person}{Marya~E Corden}, \bibinfo{person}{Konrad~P Kording}, \bibinfo{person}{David~C Mohr}, {et~al\mbox{.}}} \bibinfo{year}{2015}\natexlab{}.
\newblock \showarticletitle{Mobile phone sensor correlates of depressive symptom severity in daily-life behavior: an exploratory study}.
\newblock \bibinfo{journal}{\emph{Journal of medical Internet research}} \bibinfo{volume}{17}, \bibinfo{number}{7} (\bibinfo{year}{2015}), \bibinfo{pages}{e4273}.
\newblock


\bibitem[Sajno et~al\mbox{.}(2023)]%
        {sajno2023machine}
\bibfield{author}{\bibinfo{person}{Elena Sajno}, \bibinfo{person}{Sabrina Bartolotta}, \bibinfo{person}{Cosimo Tuena}, \bibinfo{person}{Pietro Cipresso}, \bibinfo{person}{Elisa Pedroli}, {and} \bibinfo{person}{Giuseppe Riva}.} \bibinfo{year}{2023}\natexlab{}.
\newblock \showarticletitle{Machine learning in biosignals processing for mental health: A narrative review}.
\newblock \bibinfo{journal}{\emph{Frontiers in Psychology}}  \bibinfo{volume}{13} (\bibinfo{year}{2023}), \bibinfo{pages}{1066317}.
\newblock


\bibitem[Shatte et~al\mbox{.}(2019)]%
        {shatte2019machine}
\bibfield{author}{\bibinfo{person}{Adrian~BR Shatte}, \bibinfo{person}{Delyse~M Hutchinson}, {and} \bibinfo{person}{Samantha~J Teague}.} \bibinfo{year}{2019}\natexlab{}.
\newblock \showarticletitle{Machine learning in mental health: a scoping review of methods and applications}.
\newblock \bibinfo{journal}{\emph{Psychological medicine}} \bibinfo{volume}{49}, \bibinfo{number}{9} (\bibinfo{year}{2019}), \bibinfo{pages}{1426--1448}.
\newblock


\bibitem[Stewart and Velupillai(2021)]%
        {stewart2021applied}
\bibfield{author}{\bibinfo{person}{Robert Stewart} {and} \bibinfo{person}{Sumithra Velupillai}.} \bibinfo{year}{2021}\natexlab{}.
\newblock \showarticletitle{Applied natural language processing in mental health big data}.
\newblock \bibinfo{journal}{\emph{Neuropsychopharmacology}} \bibinfo{volume}{46}, \bibinfo{number}{1} (\bibinfo{year}{2021}), \bibinfo{pages}{252}.
\newblock


\bibitem[Torous et~al\mbox{.}(2016)]%
        {torous2016new}
\bibfield{author}{\bibinfo{person}{John Torous}, \bibinfo{person}{Mathew~V Kiang}, \bibinfo{person}{Jeanette Lorme}, \bibinfo{person}{Jukka-Pekka Onnela}, {et~al\mbox{.}}} \bibinfo{year}{2016}\natexlab{}.
\newblock \showarticletitle{New tools for new research in psychiatry: a scalable and customizable platform to empower data driven smartphone research}.
\newblock \bibinfo{journal}{\emph{JMIR mental health}} \bibinfo{volume}{3}, \bibinfo{number}{2} (\bibinfo{year}{2016}), \bibinfo{pages}{e5165}.
\newblock


\bibitem[Tutun et~al\mbox{.}(2023)]%
        {tutun2023ai}
\bibfield{author}{\bibinfo{person}{Salih Tutun}, \bibinfo{person}{Marina~E Johnson}, \bibinfo{person}{Abdulaziz Ahmed}, \bibinfo{person}{Abdullah Albizri}, \bibinfo{person}{Sedat Irgil}, \bibinfo{person}{Ilker Yesilkaya}, \bibinfo{person}{Esma~Nur Ucar}, \bibinfo{person}{Tanalp Sengun}, {and} \bibinfo{person}{Antoine Harfouche}.} \bibinfo{year}{2023}\natexlab{}.
\newblock \showarticletitle{An AI-based decision support system for predicting mental health disorders}.
\newblock \bibinfo{journal}{\emph{Information Systems Frontiers}} \bibinfo{volume}{25}, \bibinfo{number}{3} (\bibinfo{year}{2023}), \bibinfo{pages}{1261--1276}.
\newblock


\bibitem[Wang et~al\mbox{.}(2014)]%
        {wang2014studentlife}
\bibfield{author}{\bibinfo{person}{Rui Wang}, \bibinfo{person}{Fanglin Chen}, \bibinfo{person}{Zhenyu Chen}, \bibinfo{person}{Tianxing Li}, \bibinfo{person}{Gabriella Harari}, \bibinfo{person}{Stefanie Tignor}, \bibinfo{person}{Xia Zhou}, \bibinfo{person}{Dror Ben-Zeev}, {and} \bibinfo{person}{Andrew~T Campbell}.} \bibinfo{year}{2014}\natexlab{}.
\newblock \showarticletitle{StudentLife: assessing mental health, academic performance and behavioral trends of college students using smartphones}. In \bibinfo{booktitle}{\emph{Proceedings of the 2014 ACM international joint conference on pervasive and ubiquitous computing}}. \bibinfo{pages}{3--14}.
\newblock


\bibitem[Wicke et~al\mbox{.}(2022)]%
        {wicke2022update}
\bibfield{author}{\bibinfo{person}{Felix~Sebastian Wicke}, \bibinfo{person}{Lina Krakau}, \bibinfo{person}{Bernd L{\"o}we}, \bibinfo{person}{Manfred~E Beutel}, {and} \bibinfo{person}{Elmar Br{\"a}hler}.} \bibinfo{year}{2022}\natexlab{}.
\newblock \showarticletitle{Update of the standardization of the Patient Health Questionnaire-4 (PHQ-4) in the general population}.
\newblock \bibinfo{journal}{\emph{Journal of Affective Disorders}}  \bibinfo{volume}{312} (\bibinfo{year}{2022}), \bibinfo{pages}{310--314}.
\newblock


\bibitem[{World Health Organization}(2024)]%
        {Theoverw93:online}
\bibfield{author}{\bibinfo{person}{{World Health Organization}}.} \bibinfo{year}{2024}\natexlab{}.
\newblock \bibinfo{title}{The Overwhelming Case for Ending Stigma and Discrimination in Mental Health}.
\newblock
\urldef\tempurl%
\url{https://www.who.int/europe/news/item/26-06-2024-the-overwhelming-case-for-ending-stigma-and-discrimination-in-mental-health}
\showURL{%
\tempurl}
\newblock
\shownote{Accessed: 2025-03-05}.


\bibitem[Xu and Shuttleworth(2024)]%
        {xu2024medical}
\bibfield{author}{\bibinfo{person}{Hanhui Xu} {and} \bibinfo{person}{Kyle Michael~James Shuttleworth}.} \bibinfo{year}{2024}\natexlab{}.
\newblock \showarticletitle{Medical artificial intelligence and the black box problem: a view based on the ethical principle of “do no harm”}.
\newblock \bibinfo{journal}{\emph{Intelligent Medicine}} \bibinfo{volume}{4}, \bibinfo{number}{1} (\bibinfo{year}{2024}), \bibinfo{pages}{52--57}.
\newblock


\bibitem[Xuan et~al\mbox{.}(2025)]%
        {xuan2025unlocking}
\bibfield{author}{\bibinfo{person}{Wei Xuan}, \bibinfo{person}{Meghna~Roy Chowdhury}, \bibinfo{person}{Yi Ding}, {and} \bibinfo{person}{Yixue Zhao}.} \bibinfo{year}{2025}\natexlab{}.
\newblock \showarticletitle{Unlocking Mental Health: Exploring College Students' Well-being through Smartphone Behaviors}.
\newblock \bibinfo{journal}{\emph{arXiv preprint arXiv:2502.08766}} (\bibinfo{year}{2025}).
\newblock


\bibitem[Yu et~al\mbox{.}(2024)]%
        {sheng_yu_careforme_2024}
\bibfield{author}{\bibinfo{person}{Sheng Yu}, \bibinfo{person}{Narjes Nourzad}, \bibinfo{person}{Randye~J. Semple}, \bibinfo{person}{Yixue Zhao}, \bibinfo{person}{Emily Zhou}, {and} \bibinfo{person}{Bhaskar Krishnamachari}.} \bibinfo{year}{2024}\natexlab{}.
\newblock \showarticletitle{CAREForMe: Contextual Multi-Armed Bandit Recommendation Framework for Mental Health}. In \bibinfo{booktitle}{\emph{Proceedings of the 11th IEEE/ACM International Conference on Mobile Software Engineering and Systems}} (Portugal) \emph{(\bibinfo{series}{MOBILESoft '24})}.
\newblock


\bibitem[Zhao et~al\mbox{.}(2024)]%
        {zhao2024digital}
\bibfield{author}{\bibinfo{person}{Yixue Zhao}, \bibinfo{person}{Tianyi Li}, {and} \bibinfo{person}{Michael Sobolev}.} \bibinfo{year}{2024}\natexlab{}.
\newblock \showarticletitle{Digital Wellbeing Redefined: Toward User-Centric Approach for Positive Social Media Engagement}. In \bibinfo{booktitle}{\emph{Proceedings of the IEEE/ACM 11th International Conference on Mobile Software Engineering and Systems}}. \bibinfo{pages}{95--98}.
\newblock


\end{thebibliography}
